%% file: arxiv.tex
\documentclass[10pt,twocolumn,letterpaper]{article}

\usepackage[pagenumbers]{cvpr} %

\input{preamble}

\usepackage[pagebackref,breaklinks,colorlinks,bookmarksdepth=3,bookmarksopen=true,bookmarksopenlevel=2,bookmarksnumbered=true]{hyperref}

\def\confName{CVPR}
\def\confYear{2027}

\title{FiatLux: Video Relighting with Lightgrams and Reciprocal Illumination Learning}

\author{
Zhiyang Guo$^{1}$\thanks{Equal contribution.} \quad
Yanni Zhang$^{1*}$\thanks{Project lead.} \quad
Hanlei Guo$^{1*}$ \quad
Haolin Lu$^{1,2}$ \quad
Alan Zhao$^{1}$ \quad
Ran Zhang$^{1\text{\Letter}}$
\\[0.5em]
$^{1}$Tencent \quad
$^{2}$University of California
\\[0.5em]
{\tt\small \{zhiyangguo, yanniizhang, hannahlguo, alantzhao\}@tencent.com,} \\
{\tt\small hal128@ucsd.edu,}
{\tt\small ranorizhang@global.tencent.com}
}

\begin{document}

\maketitle

\begingroup
\renewcommand{\thefootnote}{\Letter}
\footnotetext{Corresponding author.}
\endgroup

\input{fig/teaser}
\input{sec/0_abstract}
\input{sec/1_intro}
\input{sec/2_related}
\input{sec/3_method}
\input{sec/4_exp}
\input{sec/5_conclusion}

{
    \small
    \bibliographystyle{ieeenat_fullname}
    \bibliography{main}
}

\input{sec/X_suppl}

\end{document}

%% file: preamble.tex
\usepackage{cuted} %

\usepackage{currfile} %

\usepackage{caption} %

\usepackage{microtype}

\renewcommand{\paragraph}[1]{\vspace{.5em}\noindent\textbf{#1.}}

\usepackage{xspace}
\newcommand{\paper}{\textsc{FiatLux}\xspace}

\patchcmd{\abstract}
{\vspace*{12pt}}{\vspace*{6pt}}{}{}

\patchcmd{\endabstract}
{\vspace*{12pt}}{\vspace*{4pt}}{}{}

\usepackage{wrapfig}
\usepackage{multirow}
\usepackage{bbm}
\usepackage{bm}
\usepackage{booktabs}
\usepackage{marvosym}

\allowdisplaybreaks[4]

%% file: fig/teaser.tex
\begin{strip}
\centering
\vspace{-16mm}
\includegraphics[width=\linewidth]{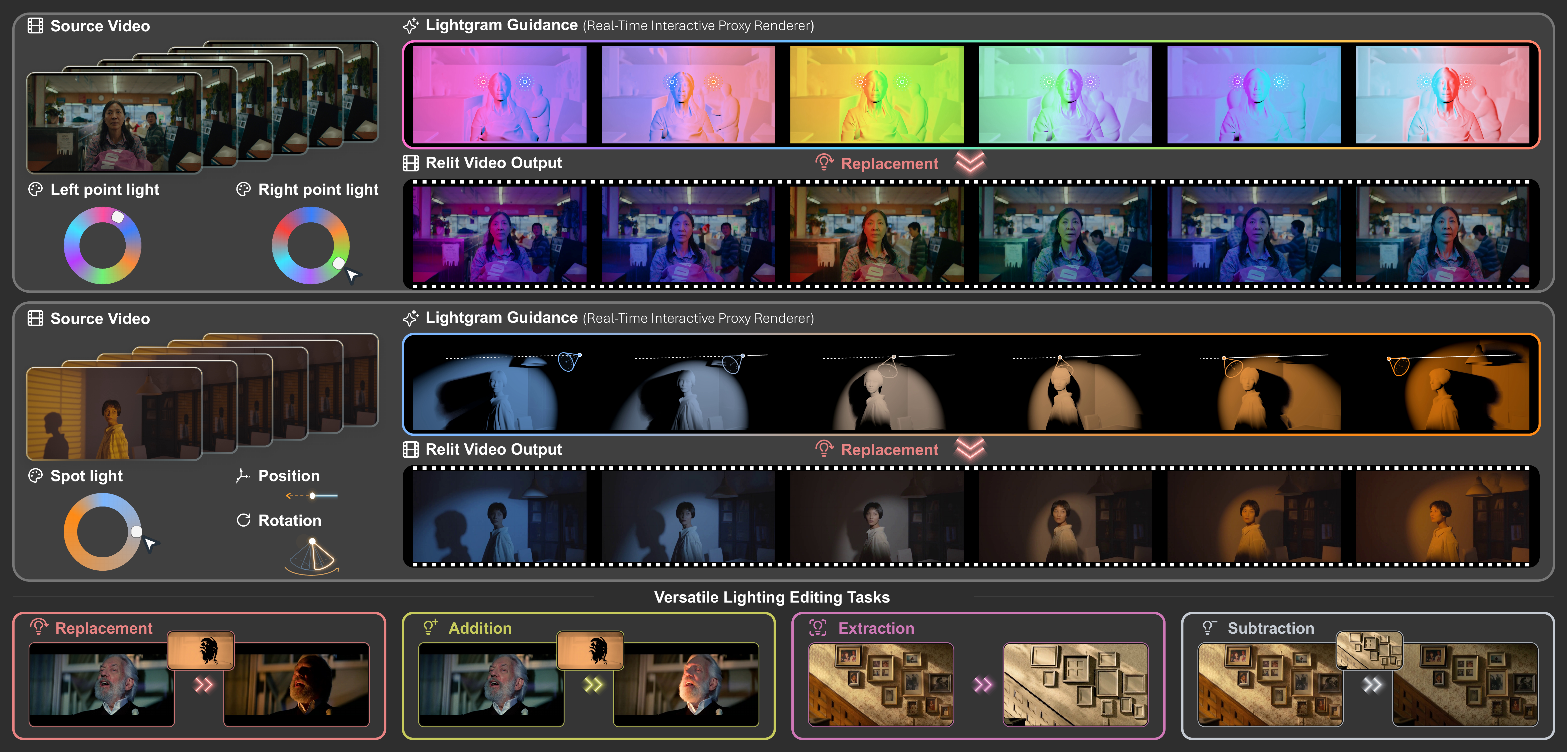}
\captionsetup{hypcap=false}
\captionof{figure}{Our interactive video relighting system \paper provides real-time lighting previews and generates realistic videos conditioned on the proposed lightgram representation. This unified framework supports flexible lighting control and versatile lighting editing tasks.}
\captionsetup{hypcap=true}
\label{fig:\currfilebase}%
\end{strip}

%% file: sec/0_abstract.tex
\begin{abstract}
Video relighting enables creators to reshape the mood and visual emphasis of captured footage.
However, existing methods lack real-time visual feedback, limit editing flexibility, and struggle with precise spatiotemporal control, while scarce real-world pairs hinder realistic generation.
To address these challenges, we present \emph{\paper}, an interactive video relighting system built around Lightgram---a pixel-aligned video of approximate direct-lighting responses rendered with a standardized material on a screen-space proxy.
By driving both real-time shader previews and generative conditioning, lightgrams enable rapid iteration across diverse light types and 3D trajectories.
Leveraging this aligned representation, we formulate relighting as in-context video generation and introduce interleaved RoPE alignment to enforce spatiotemporal consistency across input and target streams.
To train under limited real-world supervision, we propose reciprocal illumination learning, which jointly optimizes a unified model on relighting, lightgram extraction, and incremental light addition and subtraction.
Its three-stage curriculum first establishes illumination grounding on synthetic pairs, then introduces incremental editing before adapting to real-world videos using the model's own lightgram estimates.
Extensive experiments show that our method outperforms evaluated baselines while producing realistic, temporally coherent results across diverse real-world scenes.
\end{abstract}

%% file: sec/1_intro.tex
\section{Introduction}
\label{sec:intro}

Lighting shapes scene appearance, establishes mood, and guides the viewer's attention, making it central to visual storytelling.
Video relighting brings this expressive flexibility to post-production, allowing filmmakers and content creators to redesign illumination after capture and avoid costly reshoots.
In practice, lighting design is an inherently visual and iterative process: artists position light sources, observe how highlights and shadows take shape across the scene, and continuously refine the setup until achieving the intended effect.
Supporting this iterative workflow requires not only precise control over light types, colors, spatial distributions, and temporal dynamics, but also immediate visual feedback before committing to expensive video generation.

Existing generative relighting methods fall short of this interactive paradigm.
Approaches conditioned on text prompts~\cite{relightvid,lightavideo,unilumos}, reference images~\cite{iclight,relightvid}, or environment maps~\cite{unirelight,diffusionrenderer,relitlive} convey global ambiance but offer limited control over the precise placement and spatial extent of local lighting effects.
To enable more direct spatial control, recent works encode 3D point lights via multi-plane light images (MPLIs)~\cite{relightmaster,livelight} on fixed depth planes lacking direct scene correspondence, or rely on 2D trajectories~\cite{lightctrl} without explicit depth control, whereas shading-cue methods target single images or distant illumination~\cite{dilightnet,jointlight}.

Moreover, these frameworks typically lock users into a single editing mode---either replacing captured illumination or adding new lights~\cite{livelight}---rather than supporting both capabilities, and even light removal, together.
Beyond these constraints, without real-time visual feedback on how specified lights affect the scene, users must rely on costly generation passes to assess outcomes.
Together, these limitations prevent a workflow that is \emph{fast} to iterate, \emph{flexible} across light types and editing tasks, and \emph{spatiotemporally controllable}.

Our key insight is that these goals can be achieved by bridging interactive preview and generative control through a shared lighting representation.
To realize this, we present \paper (Latin for ``let there be light''), an interactive video relighting system built around \emph{Lightgram}.
A lightgram is a pixel-aligned video of approximate direct-lighting responses, rendered with a standardized material on a screen-space proxy constructed from estimated depth and normal maps.

This design establishes a clean division of labor: the lightgram specifies \emph{where light falls and how it evolves}, while the source video supplies the underlying \emph{scene content and texture}.
By encoding these rendered responses in a unified format, lightgrams accommodate point, spot, directional, and area lights with varying counts and dynamic 3D trajectories.
Users can interactively configure lights in our real-time shader and preview coarse effects on the source video, passing the corresponding lightgram directly to guide generation.

Since the lightgram representation shares the exact same pixel-aligned format as RGB video, the relighting task can be directly formulated as in-context generation by adapting a pretrained video backbone~\cite{h3}.
To maintain consistency across aligned input and output streams, we propose \emph{Interleaved RoPE Alignment}, which assigns corresponding tokens the same base spatiotemporal coordinates with small, role-specific temporal offsets.
This alignment provides a strong prior for the attention mechanism, guiding the model to faithfully combine textured appearance with illumination.

Beyond relighting, this shared representation space also allows the same backbone to learn inverse illumination estimation, addressing a core training bottleneck: real-world videos with identical motion under varying illumination are difficult to capture, whereas synthetic rendering suffers from an appearance domain gap.
We bridge this gap with \emph{Reciprocal Illumination Learning}, a multi-task strategy that jointly trains a unified model on the core relighting task alongside lightgram extraction and incremental light addition and subtraction.
A three-stage curriculum first develops illumination grounding on synthetic relighting pairs, then incorporates incremental editing, and finally adapts the model to real-world videos by leveraging its learned extraction ability to supply lightgram conditions for unannotated footage.

Extensive experiments demonstrate that our method substantially outperforms evaluated open-source baselines, while achieving qualitative results comparable to or better than advanced commercial models.

In summary, our contributions are as follows:
\begin{itemize}
\item We present \paper, an interactive video relighting system built around \emph{Lightgram}, a pixel-aligned illumination representation that connects real-time lighting previews with generative control across diverse light types, attributes, and dynamics.
\item By introducing \emph{Interleaved RoPE Alignment} and \emph{Reciprocal Illumination Learning}, we develop a unified in-context generative model that bridges video relighting with incremental lighting edits and illumination extraction, enabling effective adaptation to unannotated real-world videos.
\item Extensive experiments demonstrate that our method faithfully aligns with specified lighting intents while producing realistic, temporally coherent results across diverse real-world scenarios.
\end{itemize}

%% file: sec/2_related.tex
\section{Related Work}
\label{sec:related}

\begin{figure*}[t]
  \centering
  \vspace{-3mm}
    \includegraphics[width=1.0\linewidth]{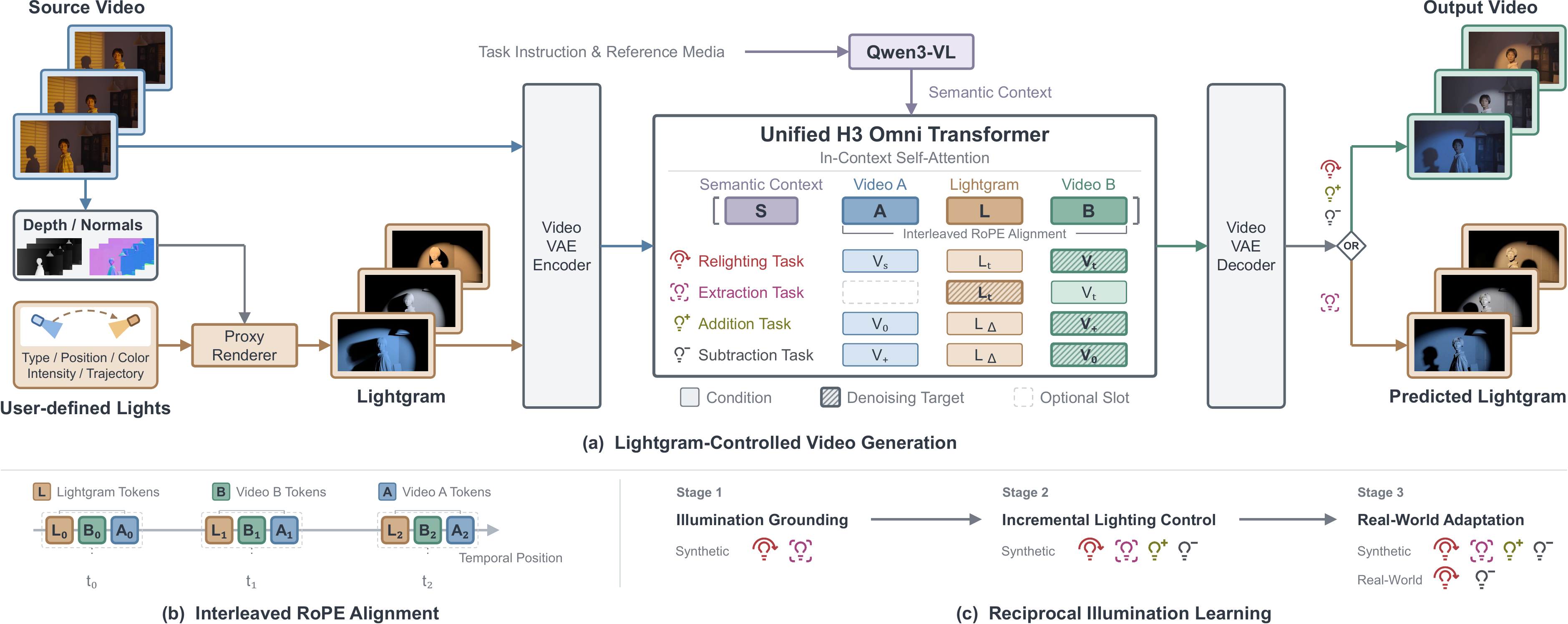}
  \caption{\textbf{Method overview.}
  \textbf{(a)} We render lightgrams from estimated depth and normals under user-defined lights. A shared in-context Transformer processes the context latents with task-specific denoising targets.
  \textbf{(b)} Interleaved RoPE Alignment groups corresponding tokens with fixed, stream-specific temporal offsets.
  \textbf{(c)} Reciprocal Illumination Learning progresses from synthetic relighting and illumination extraction to addition and subtraction, then incorporates real-world adaptation tasks.
}
  \label{fig:pipeline}
\end{figure*}

\begin{figure}[t]
  \centering
  \includegraphics[width=0.85\linewidth]{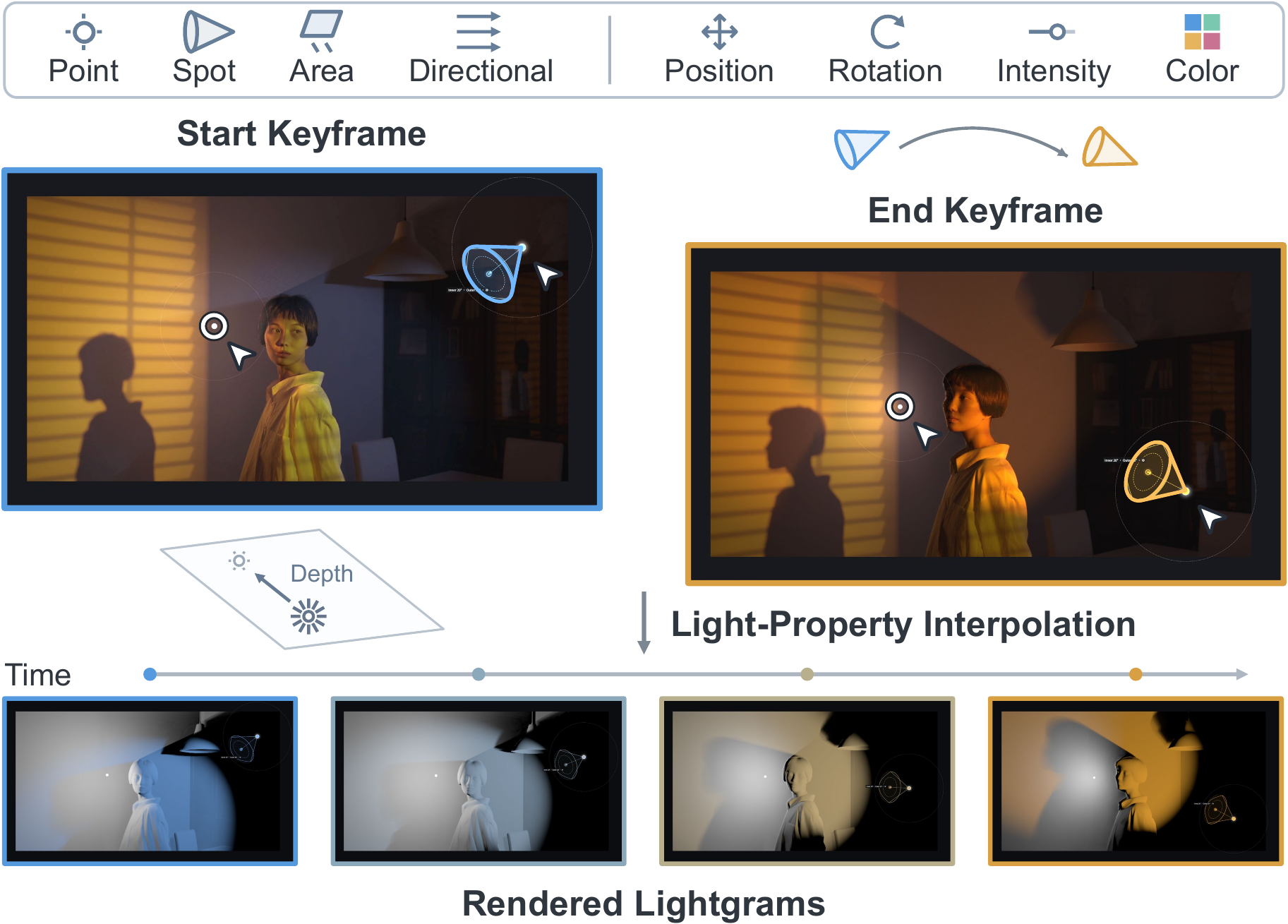}
  \caption{\textbf{Interactive lighting control.}
  Users can adjust lighting configurations with real-time effect previews on the source video.
  The corresponding lightgram can then condition video generation.}
  \label{fig:shader}
\end{figure}

\subsection{Image and Video Relighting}

Relighting has long been studied in computer graphics and computer vision.
Earlier image-based approaches rely on estimated portrait geometry and reflectance for re-rendering~\cite{totalrelighting,switchlight,lightpainter}, while recent diffusion models directly synthesize relit scenes without explicit decomposition~\cite{synthlight,iclight,neuralgaffer}.
Research on controllability has also explored editable intrinsic channels~\cite{rgbx,intrinsicedit}, geometry-derived shading cues rendered with reference materials~\cite{dilightnet,lightit}, or local guidance through scribbles, masks, and lighting attributes~\cite{scribblelight,lightlab,tokenlight}.

Video relighting further introduces the challenge of temporal consistency.
Existing works address this through conditional video generation~\cite{relightvid,luxpostfacto,lumen,unilumos}, training-free combinations of pretrained models~\cite{lightavideo,lightctrl}, or keyframe edit propagation~\cite{vrgbx,idv2v}.
In parallel, studies of lighting control explore intrinsic properties and rendered lighting cues~\cite{diffusionrenderer,unirelight,jointlight,relitlive}, as well as multi-plane representations of dynamic point lights~\cite{relightmaster,livelight}.
In this work, to combine explicit lighting control with realistic appearance, we condition video generation on a pixel-aligned lighting representation.

Notably, concurrent work JointLight~\cite{jointlight} jointly denoises RGB video and irradiance for human-centric relighting under environment maps, using cues rendered from reconstructed meshes.
We instead render a screen-space shading proxy, supporting diverse light types and trajectories across general scenes, while linking real-time shader previews to generation conditions.
Our training strategy further enables versatile lighting controls within a unified framework, a capability generally lacking in existing methods.

\subsection{Controllable Video Generation}

Modern video generation methods typically extend text guidance with structural conditions~\cite{controlvideo}, camera trajectories~\cite{cameractrl, gen3c}, and motion or performance references~\cite{gowiththeflow,wananimate}.
To support diverse generation and editing tasks, VACE~\cite{vace} uses a shared conditioning interface and context adapters.
In-context frameworks such as UNIC~\cite{unic} and MiniMax H3~\cite{h3} unify these tasks by jointly processing reference inputs and generation tokens within a shared Transformer.
This design offers flexible conditioning and expressive cross-modal interactions, but maintaining fidelity to multiple references remains challenging.
Building upon H3, we introduce interleaved RoPE alignment across streams to achieve better content and temporal consistency between generated videos and the conditions.

%% file: sec/3_method.tex
\section{Method}
\label{sec:method}

As shown in \cref{fig:pipeline}, we represent user-specified illumination with lightgrams (\cref{sec:lightgram}) and use an aligned in-context model to relight videos (\cref{sec:relighting_model}).
Our data engine (\cref{sec:data_engine}) provides synthetic and real-world supervision for reciprocal illumination learning (\cref{sec:reciprocal_learning}), which jointly trains the core relighting with complementary tasks.

\begin{figure*}[t]
  \centering
  \vspace{-5mm}
  \includegraphics[width=0.95\linewidth]{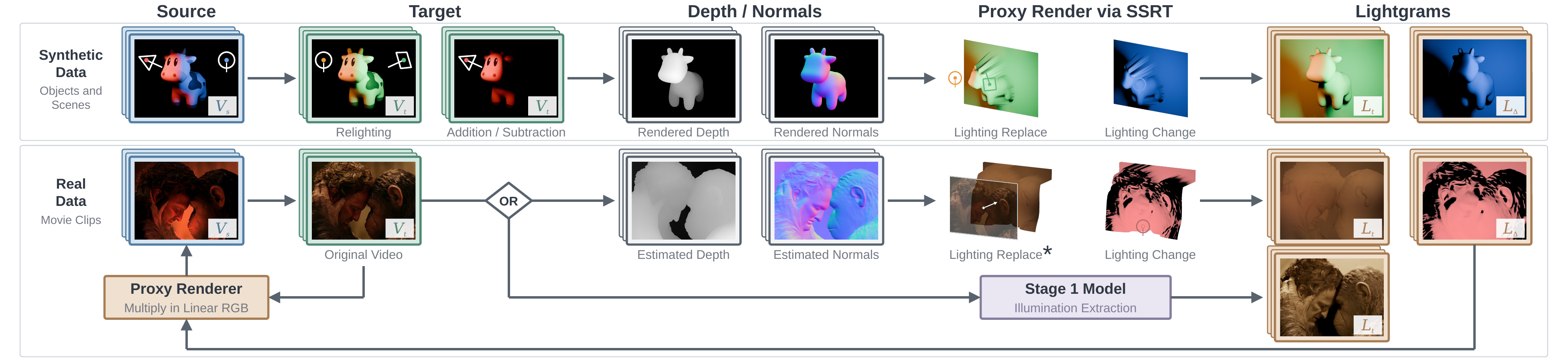}
  \caption{\textbf{Data curation pipeline.} \textit{Top:} Rendering engines produce synthetic pairs $(V_s,V_t)$; our proxy shader renders $L_t$ and $L_\Delta$ via SSRT on screen-space proxies. \textit{Bottom:} Original footage serves as $V_t$. We synthesize perturbed $V_s$ by apply shader-rendered $L_\Delta$ on $V_t$. $L_t$ is extracted by the Stage~1 model or rendered with the proxy shader. $*$ denotes area light textured with the original frames.}
  \label{fig:data_curation}
\end{figure*}

\subsection{Lightgrams for Interactive Control}
\label{sec:lightgram}

To support interactive lighting control, we represent lighting configurations as \textit{Lightgrams}, videos of standardized direct-illumination responses aligned with the source video in space and time.
This illumination representation encodes light color, intensity, and spatiotemporal distribution for varying numbers of point, spot, directional, or area lights.

For each frame $k=1,\ldots,T$, monocularly estimated depth and normal maps define a screen-space shading proxy $G^{(k)}$.
Given the lighting configuration $\Lambda^{(k)}$, we render
\begin{equation}
  L^{(k)} = \mathcal T\!\left(\mathcal R_{\mathrm{proxy}}(G^{(k)},\Lambda^{(k)})\right) \in [0,1]^{H\times W\times 3},
  \label{eq:lightgram}
\end{equation}
where $\mathcal R_{\mathrm{proxy}}$ computes lighting responses by directly querying the maps in $G^{(k)}$, assuming a standardized material and approximating visibility via screen-space ray tracing (SSRT).
$\mathcal T$ maps the linear RGB responses to $[0,1]$, and $H\times W$ denotes the spatial resolution.
Although lightgram frames are two-dimensional, changes in shading and shadows can still encode light motion along the camera's depth axis.

We implement this rendering process in a real-time interactive proxy shader (\cref{fig:shader}), allowing users to design dynamic lighting with immediate visual feedback.
To preview light placement and color effects on the source video, we multiply source frames by the lightgram in linear RGB and convert back to sRGB. This coarse preview remains influenced by the captured illumination and serves as a qualitative guide for lighting design.
The lightgrams then condition generation on direct illumination, guiding the model to synthesize the selected lighting edits.

\subsection{Aligned In-Context Relighting}
\label{sec:relighting_model}

To generate relit videos from the rendered lighting cues, we condition an H3 Omni Transformer~\cite{h3} on the source video and lightgram.
A shared video VAE encodes the source, lightgram, and target videos as $Z_s$, $Z_l$, and $Z_t$, while Qwen3-VL~\cite{qwen3vl} encodes task prompts and reference media into semantic context $S$.
The Transformer processes $[S;Z_s;Z_l;Z_t^{\tau}]$ through full self-attention, combining source appearance and lighting guidance to denoise the target latent $Z_t^{\tau}$ at flow time $\tau$.

To maintain spatiotemporal correspondence across these streams, we introduce \emph{Interleaved RoPE Alignment} (\cref{fig:pipeline}).
The three streams share spatial coordinates $(u,v)$ and the backbone's native RoPE~\cite{roformer} time coordinate $t_k$ at latent frame $k$.
For each stream slot $r$, we add a distinct temporal offset $\delta_r$ fixed across the sequence:
\begin{equation}
  \begin{gathered}
    P_r(k,u,v)=(t_k+\delta_r,u,v),\\
    \max_r\delta_r-\min_r\delta_r<\Delta_{\min},
  \end{gathered}
  \label{eq:alignment}
\end{equation}
where $\Delta_{\min}:=\min_k(t_{k+1}-t_k)$ is the minimum spacing between consecutive temporal coordinates.
This layout groups corresponding tokens while distinguishing stream roles, and the offset bound keeps adjacent frame groups separate.
It provides attention with a correspondence prior for combining source appearance and lighting guidance.

\subsection{Data Engine}
\label{sec:data_engine}
To train our model, we propose a data engine that constructs source--target video pairs $(V_s,V_t)$ with shared scene content and motion but different illumination, together with aligned lightgrams (\cref{fig:data_curation}).
Here, $L_t$ represents the target illumination, and $L_\Delta$ represents an incremental lighting component.

For synthetic data, we use rendering engines to produce object-level and scene-level pairs under different or incrementally modified illumination, with flat and neutral source lighting to reduce material–lighting ambiguity.
Ground-truth depth, normals, and lighting parameters drive the proxy shader (\cref{sec:lightgram}) to render $L_t$ for relighting and $L_\Delta$ for light addition and subtraction. 

For real-world data, paired lighting captures are rarely available, so we use the original footage as the target $V_t$ and synthesize a perturbed source $V_s$ from it. We obtain $V_s$ by applying a sampled incremental lightgram $L_\Delta$ on $V_t$. For subtraction, we directly use the known perturbation $L_\Delta$ as the condition.
For relighting, we estimate the original lighting $L_t$ from $V_t$ with our extraction model (\cref{sec:reciprocal_learning}), or otherwise approximate it with dynamic area lights textured with the original frames.

An off-the-shelf MLLM~\cite{google2026gemini38flash} separately captions appearance and lightgram illumination to derive task-specific instructions. We discard samples with negligible appearance changes or excessively dark renders.

\subsection{Reciprocal Illumination Learning}
\label{sec:reciprocal_learning}

To translate coarse lightgrams into realistic lighting effects, the model must learn how illumination changes scene appearance.
We therefore introduce \emph{Reciprocal Illumination Learning}, a unified framework that consumes paired data to jointly train relighting with complementary tasks for flexible lighting control and better illumination understanding.

The core relighting task $(V_s,L_t)\mapsto V_t$ replaces the source illumination with that specified by $L_t$.
We pair this task with lightgram extraction $V_t\mapsto L_t$ to learn the relationship between illumination and appearance in both directions.
For incremental lighting control, we additionally pair light addition $(V_0,L_\Delta)\mapsto V_+$ with subtraction $(V_+,L_\Delta)\mapsto V_0$.
Here, $V_0$ and $V_+$ depict the same scene before and after adding the lighting component specified by $L_\Delta$.

All four tasks share the context sequence layout and RoPE role offsets described in \cref{sec:relighting_model}.
The available conditioning blocks and the target block selected for denoising vary by task.
To specify the intended lighting operation within this unified model, we add zero-initialized, learnable task embeddings to its AdaLN-style modulation parameters.
For a sampled task $q$ with clean target latent $Z$ and conditions $C_q$, we use a shared conditional flow-matching objective~\cite{flowmatching}:
\begin{equation}
  \mathcal L_q = \mathbb E\!\left[w(\tau)\left\|v_\theta(Z^{\tau},C_q,\tau,q)-(Z-\epsilon)\right\|_2^2\right],
  \label{eq:flow_matching}
\end{equation}
where $Z^{\tau}=\tau Z+(1-\tau)\epsilon$, $\epsilon\sim\mathcal N(0,I)$, and $w(\tau)$ weights the loss at flow time $\tau\in[0,1]$.
The velocity predictor $v_\theta$ uses conditions $C_q$, comprising semantic context $S$ and the available conditioning latents.
The clean VAE latent $Z$ represents a lightgram video for extraction and an RGB video for the other tasks.

For effective training, we introduce these tasks progressively through three stages.
Specifically, Stage~1 grounds illumination understanding through relighting and extraction on synthetic pairs.
Building on this grounding, Stage~2 introduces addition and subtraction on synthetic incremental pairs to develop control over individual lighting components.

To adapt the model to the real domain, Stage~3 adds real-world photometric restoration pairs for relighting and subtraction while retaining synthetic supervision for all four tasks.
Both real-world tasks take the perturbed video $V_s$ as input and recover the original footage $V_t$.
Relighting is conditioned on the original illumination $L_t$ extracted by our Stage~1 model, allowing training without explicit lighting annotations.
In turn, subtraction is conditioned on the applied perturbation $L_\Delta$, adapting the model to real-world appearance and proxy-rendered cues.
Given the approximate nature of these illumination perturbations (\cref{sec:lightgram}), we prioritize relighting over subtraction on real data. More task-balancing details are provided in the supplementary material.

%% file: sec/4_exp.tex
\section{Experiments}
\label{sec:exp}

\begin{figure*}[t]
  \centering
  \vspace{-5mm}
  \includegraphics[width=0.95\linewidth]{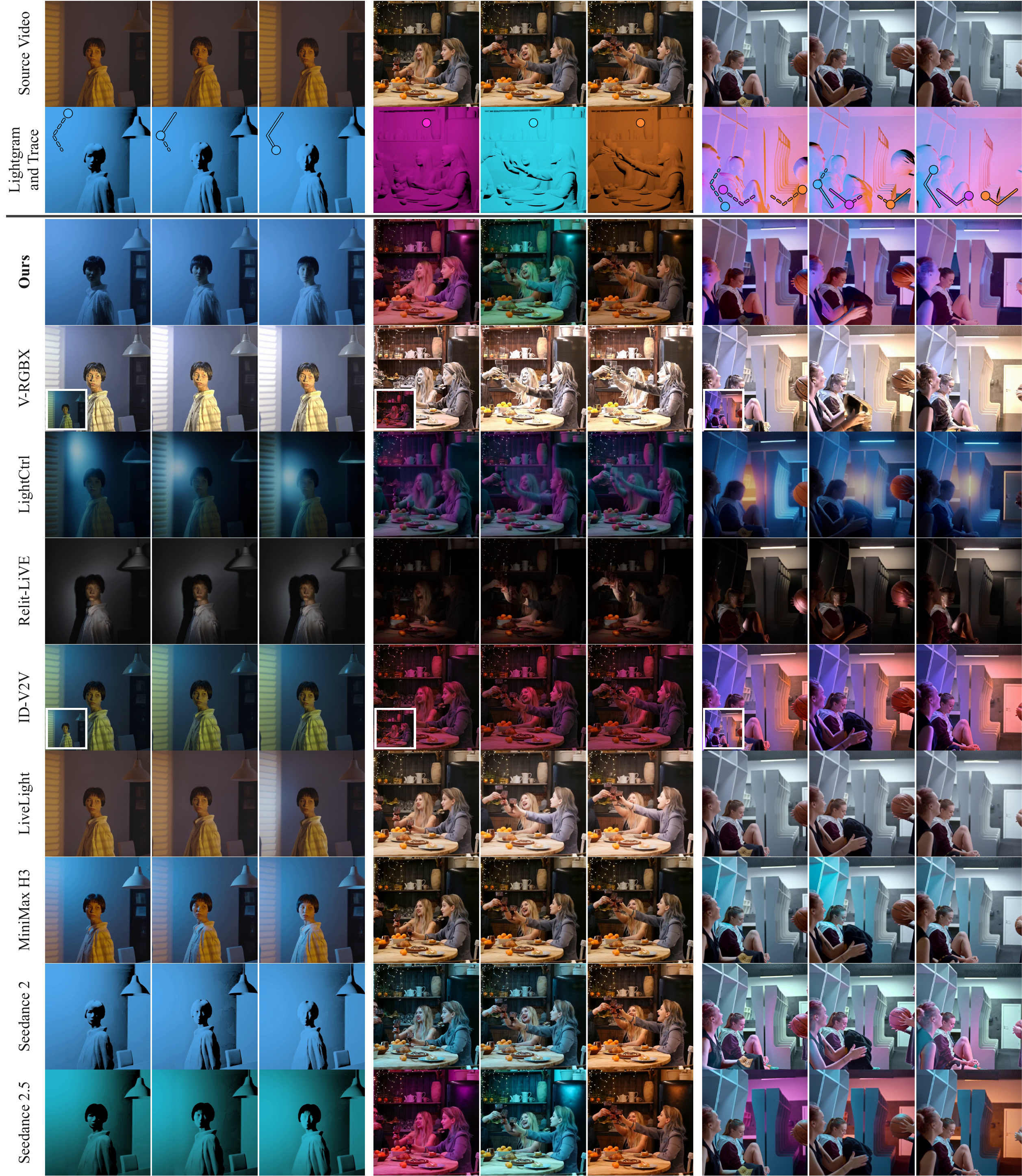}
  \caption{
  \textbf{Qualitative comparison.}
  Visual comparison of our method with baseline video relighting methods.
  We use point lights to align with the baselines' lighting representations for a fair comparison.
  V-RGBX~\cite{vrgbx} and ID-V2V~\cite{idv2v} are shown under static lighting, as they take only a relit first frame as input.
  }
  \label{fig:qualitative_baselines}
\end{figure*}

\subsection{Experimental Settings}
\label{sec:exp_settings}

\paragraph{Dataset}
For evaluation, we construct a synthetic test set using the rendering pipeline described in \cref{sec:data_engine}. The test set comprises both object-level and scene-level videos. Object assets are selected from ABO~\cite{abo} and GSO~\cite{gso}, while scene-level videos are rendered from Unreal Engine~\cite{ue}. 
We also curate a set of in-the-wild videos from online sources, including Pexels~\cite{pexels}, Mixkit~\cite{mixkit}, and publicly available films. We further construct a pool of target lighting conditions. Every method relights each video with the same lighting conditions sampled from this pool.

\paragraph{Baselines}
We compare against five open-source video relighting methods with different illumination control interfaces.
Relit-LiVE~\cite{relitlive} uses global per-frame environment maps.
LightCtrl~\cite{lightctrl} uses a 2D user-defined light trajectory with a text prompt.
LiveLight~\cite{livelight} uses depth-aware Multi-Plane Light Irradiance (MPLI) maps.
V-RGBX~\cite{vrgbx} and ID-V2V~\cite{idv2v} are keyframe-based methods that propagate edited frames to the full video.
For those methods, we use the ground-truth relit keyframes on synthetic data and relit frames generated by Nano Banana~\cite{nanobanana} on real data.
We further evaluate general-purpose video editing models, including our backbone MiniMax H3~\cite{h3} and the closed-source Seedance 2.0~\cite{seedance2} and 2.5~\cite{seedance25}.
We provide each with the same source video and lightgram, following its recommended prompt format to specify the relighting task.

\paragraph{Evaluation metrics}
To assess relighting accuracy, we compare generated videos with ground-truth targets using PSNR, SSIM~\cite{ssim}, and LPIPS~\cite{lpips}.
To evaluate the visual quality and temporal stability of the relit videos, we adopt four metrics from VBench~\cite{vbench}: Aesthetic Quality, Subject Consistency, Background Consistency, and Motion Smoothness. 

We introduce a new extraction-based protocol for evaluating lighting accuracy.
Using our frozen final model, we extract lightgrams from baselines' output and compare them with the target lightgrams.
We report median RGB angular error~\cite{finlayson2017reproduction} for light-color agreement and mean SSIM~\cite{ssim} for luminance and spatial structure.
Extractions from ground-truth RGB videos provide an empirical reference.

\paragraph{Implementation details}
For the video generation backbone, we finetune MiniMax H3~\cite{h3} using LoRA~\cite{lora} with rank 256, jointly training the adapters and our task-specific modulation embeddings.
We process clips of up to 158 frames, preserving aspect ratio within a pixel budget of $768\times1344$.
Further details are provided in the supplementary material.

\subsection{Comparison Results}
\label{sec:compare}

\paragraph{Quantitative comparison}
\Cref{tab:quantitative} reports results on the synthetic test set under two task settings: illumination replacement and addition.
Our empirical tests show that LiveLight~\cite{livelight} retains the source illumination and adds the target lighting on top, rather than replacing the original lighting.
For a fair comparison, we therefore evaluate LiveLight only on light addition and compare the remaining baselines under replacement.
In both settings, our method outperforms all baselines by a clear margin.
Beyond RGB similarity, we further assess lighting accuracy in \cref{tab:lighting_accuracy}.
Our method achieves the lowest median RGB angular error and the highest lightgram SSIM, approaching the GT-extracted reference.
This indicates closer agreement with the requested light colors and spatially varying lighting responses.

On real-world videos, where ground truth is unavailable, we report VBench metrics in \cref{tab:vbench_video}. Our method achieves the best results on most metrics, indicating good visual quality and temporally stable relighting.
LiveLight scores slightly higher than ours on Subject Consistency, which we attribute to its tendency to preserve the original video content.

\input{table/syn_comp}

\input{table/lighting_accuracy}

\input{table/real_comp}

\begin{figure*}[t]
    \centering
    \vspace{-5mm}
    \includegraphics[width=1.0\linewidth]{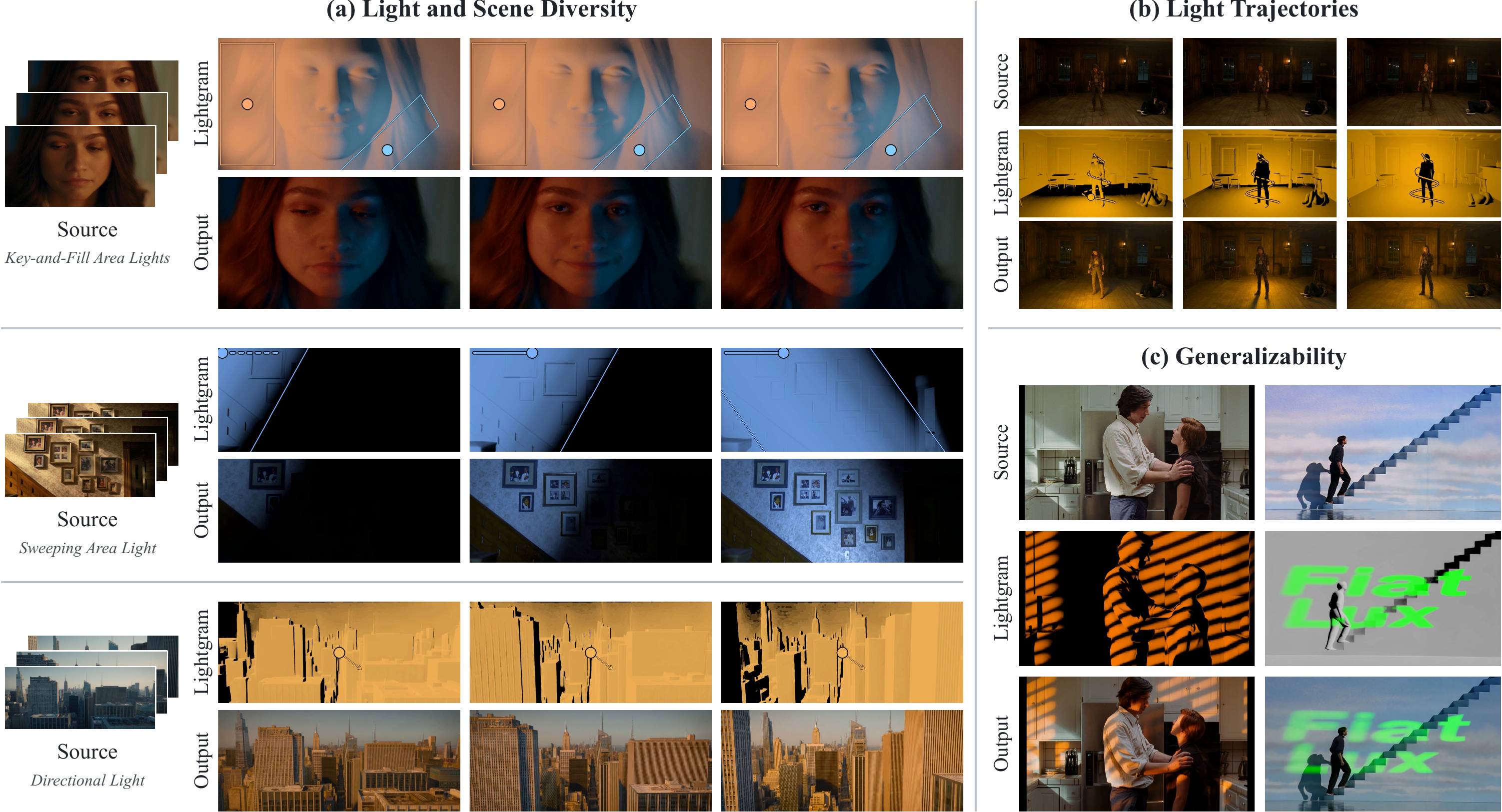}
    \caption{\textbf{Additional relighting results.} 
    (a) Close-up portraits under key-and-fill area lights with inferred eye highlights, interiors with sweeping shadows, and outdoor scenes.
    (b) A tree-like light spiral weaving around the subject.
    (c) Generalization to textured area lights.}
    \label{fig:more_vis}
\end{figure*}

\begin{figure*}[t]
  \centering
  \vspace{-2mm}
  \includegraphics[width=1.0\linewidth]{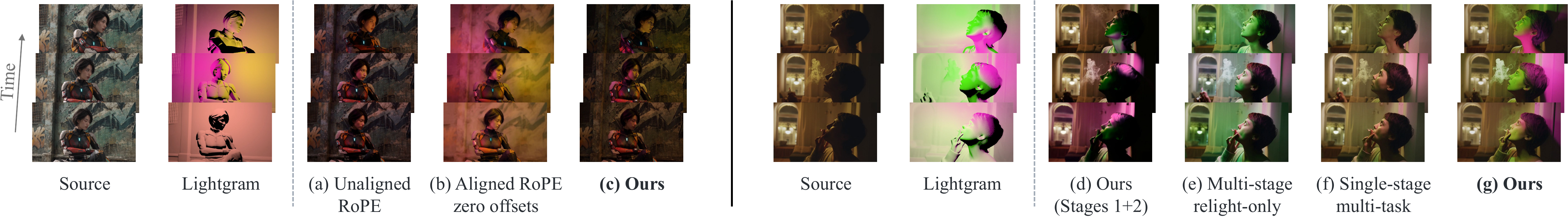}
  \caption{\textbf{Qualitative ablation results.} The variant labels correspond one-to-one with those in \cref{tab:ablation}.}
  \label{fig:ablation}
\end{figure*}

\paragraph{Qualitative comparison}
\Cref{fig:qualitative_baselines} presents qualitative comparisons on real-world videos.
Our method faithfully follows the target lighting while preserving the original scene appearance, whereas the baselines exhibit noticeable limitations.
LightCtrl~\cite{lightctrl} produces inaccurate light--shadow relationships, as its 2D trajectory does not account for the depth of the light source; moreover, its text prompt provides only coarse control over light color.
Relit-LiVE~\cite{relitlive} and V-RGBX~\cite{vrgbx} rely on estimated intrinsic properties, so errors in these estimates propagate to the relit videos and degrade fidelity.
LiveLight~\cite{livelight} responds only weakly to the target lights, leaving the original illumination largely unchanged.
ID-V2V~\cite{idv2v} generates visually plausible videos, but its results are bounded by the quality of the edited first frame and cannot correct lighting errors introduced by the image editing model.
Moreover, keyframe-based methods~\cite{vrgbx,idv2v} are not well suited to time-varying lighting.

Compared with general-purpose video editing models, our method replaces illumination more reliably and follows the lightgram more closely.
Pretrained H3 sometimes adds plausible light but poorly follows the lightgram and often misaligns temporally with the source.
Seedance 2.0 adds light better but often blends the source with the lightgram.
Seedance 2.5 produces more convincing lighting effects yet still shows similar blending or misplaced, miscolored lights.
Even when these methods succeed, ours matches or exceeds their visual quality.

\subsection{More Visualizations}
\label{sec:more_vis}

\Cref{fig:more_vis} demonstrates the versatility of our method in terms of scene diversity, complex lighting motion, and generalizability.
Across scene types (a), our method relights close-up portraits with matching amber and blue highlights in the eyes, while achieving faithful lighting control in interiors with sweeping light–shadow transitions and expansive outdoor scenes.
For light trajectories (b), a light spiraling around the subject at varying depths produces smooth transitions between frontal and rim lighting, with plausible floor shadows.
For generalizability (c), our method reproduces window-blind shadows and text cast by textured area lights, which are unseen during training, suggesting generalization to certain out-of-distribution illumination.

\input{table/ablation}

\subsection{Ablation Study}
\label{sec:ablation}

We report quantitative and qualitative ablation results in \cref{tab:ablation,fig:ablation}.
All results use the same test set as \cref{tab:quantitative}.
Additional discussions are provided in the supplementary.

\paragraph{Positional alignment}
With the Stage~1 training recipe fixed, aligned RoPE with role-specific offsets (\textit{c}) improves all the metrics over (\textit{a}) and (\textit{b}) (\cref{tab:ablation}).
These gains accompany better temporal correspondence and content preservation in the synthetic examples of \cref{fig:ablation}, where (\textit{a}) exhibits misalignment and (\textit{b}) mixes lightgram patterns with RGB appearance.

\paragraph{Complementary illumination tasks}
As shown in \cref{tab:ablation}, Stage~2 improves synthetic relighting from (\textit{c}) to (\textit{d}), and the full model (\textit{g}) also outperforms relight-only training (\textit{e}).
In the real-world example (\cref{fig:ablation}), (\textit{e}) produces lighting that deviates from the intended illumination, suggesting limited understanding of the lightgram condition.
The closer match in (\textit{g}) supports the role of complementary tasks in improving lightgram understanding and lighting adherence.

\paragraph{Real-world adaptation}
On real videos, the synthetic-only Stages~1+2 model (\textit{d}) produces results with a pronounced computer graphics rendering style (\cref{fig:ablation}).
Real-world adaptation (\textit{g}) improves the naturalness of facial shading and adherence to the colored lightgram, while largely preserving synthetic relighting fidelity (\cref{tab:ablation}).

\paragraph{Staged training}
Single-stage mixed training (\textit{f}) yields weaker performance on the synthetic test set (\cref{tab:ablation}) and flatter, less colorful lighting on real videos (\cref{fig:ablation}).
The full curriculum (\textit{g}) produces stronger, spatially distinct color responses, supporting the value of establishing diverse synthetic lighting capabilities before real-world adaptation.

%% file: table/syn_comp.tex
\begin{table}[t]
\vspace{-3mm}
\caption{
\textbf{Quantitative comparison} evaluated on the synthetic test set.
Best and second-best results are shown in \textbf{bold} and \underline{underlined}, respectively.
$\dagger$ V-RGBX is provided with the GT relit frames.
}
\label{tab:quantitative}
\centering
\setlength{\tabcolsep}{3pt}
\resizebox{1.0\linewidth}{!}{%
\begin{tabular}{@{}clccc@{}}
    \toprule
    Task & Method & PSNR $\uparrow$ & SSIM $\uparrow$ & LPIPS $\downarrow$ \\
    \midrule
    \multirow{5}{*}{\shortstack{Relighting\\(Replacement)}}
    & V-RGBX$^\dagger$~\cite{vrgbx}
    & 11.2011 & 0.4375 & 0.3967 \\
    & LightCtrl~\cite{lightctrl}
    & 12.6060 & 0.3913 & 0.5336 \\
    & Relit-LiVE~\cite{relitlive}
    & 14.8449 & 0.4778 & 0.4065 \\
    & MiniMax H3~\cite{h3}
    & \underline{15.3069} & \underline{0.5093} & \underline{0.3633} \\
    & \textbf{Ours}
    & \textbf{19.2039} & \textbf{0.6545} & \textbf{0.2245} \\
    \midrule
    \multirow{3}{*}{\shortstack{Light\\Addition}}
    & LiveLight~\cite{livelight}
    & \underline{14.4382} & \underline{0.6465} & \underline{0.3189} \\
    & MiniMax H3~\cite{h3}
    & 12.7784 & 0.5590 & 0.3740 \\
    & \textbf{Ours}
    & \textbf{21.4008} & \textbf{0.7615} & \textbf{0.1578} \\
    \bottomrule
\end{tabular}%
}
\end{table}

%% file: table/lighting_accuracy.tex
\begin{table}[t]
\vspace{-3mm}
    \centering
    \setlength{\tabcolsep}{3pt}
    \small
    \caption{\textbf{Extraction-based lighting accuracy} evaluated on the synthetic test set. Best and second-best results are shown in \textbf{bold} and \underline{underlined}, respectively (excluding the GT-extracted reference).}
    \label{tab:lighting_accuracy}
    \begin{tabular}{lcc}
        \toprule
        Method & Median RGB AE ($^\circ$)~$\downarrow$ & SSIM~$~\uparrow$ \\
        \midrule
        V-RGBX~\cite{vrgbx} & 8.265 & \underline{0.6116} \\
        LightCtrl~\cite{lightctrl} & 8.202 & 0.5365 \\
        Relit-LiVE~\cite{relitlive} & 8.926 & 0.5423 \\
        MiniMax H3~\cite{h3} & \underline{7.889} & 0.5912 \\
        \midrule
        \textbf{Ours} & \textbf{5.409} & \textbf{0.7135} \\
        \midrule
        \textcolor{black!60}{GT-extracted (for reference)} & \textcolor{black!60}{5.229} & \textcolor{black!60}{0.7404} \\
        \bottomrule
    \end{tabular}
\end{table}

%% file: table/real_comp.tex
\begin{table}[t]
    \centering
    \setlength{\tabcolsep}{3pt}
    \small
    \caption{\textbf{VBench evaluation}.
    Higher values are better. Best and second-best results are shown in \textbf{bold} and \underline{underlined}, respectively.}
    \label{tab:vbench_video}
        \begin{tabular}{@{}lcccc@{}}
            \toprule
            Method
            & \begin{tabular}[c]{@{}c@{}}Aesthetic\\Quality\end{tabular}\,$\uparrow$
            & \begin{tabular}[c]{@{}c@{}}Subj.\\Consist.\end{tabular}\,$\uparrow$
            & \begin{tabular}[c]{@{}c@{}}Bg.\\Consist.\end{tabular}\,$\uparrow$
            & \begin{tabular}[c]{@{}c@{}}Motion\\Smooth.\end{tabular}\,$\uparrow$ \\
            \midrule
            V-RGBX~\cite{vrgbx}
            & \underline{0.5999} & 0.9363 & 0.9405 & 0.9786 \\
            LightCtrl~\cite{lightctrl}
            & 0.5620 & 0.9425 & 0.9519 & 0.9874 \\
            Relit-LiVE~\cite{relitlive}
            & 0.5410 & 0.9467 & \underline{0.9565} & \underline{0.9903} \\
            ID-V2V~\cite{idv2v}
            & 0.5923 & 0.9438 & 0.9487 & 0.9893 \\
            LiveLight~\cite{livelight}
            & 0.5833 & \textbf{0.9497} & 0.9557 & 0.9887 \\
            MiniMax H3~\cite{h3}
            & 0.5830 & 0.9408 & 0.9485 & 0.9890 \\
            \midrule
            \textbf{Ours}
            & \textbf{0.6026} & \underline{0.9492}
            & \textbf{0.9597} & \textbf{0.9908} \\
            \bottomrule
        \end{tabular}
\end{table}

%% file: table/ablation.tex
\begin{table}[t]
\vspace{-2mm}
  \caption{
    \textbf{Quantitative ablation results} on the synthetic test set for the relighting task.
    Best and second-best results are shown in \textbf{bold} and \underline{underlined}, respectively.
    Note that (\textit{a})--(\textit{c}) compare RoPE variants under synthetic-only Stage~1 training, while (\textit{c})--(\textit{g}) all use the proposed RoPE alignment with role-specific offsets.
  }
  \label{tab:ablation}
  \centering
  \setlength{\tabcolsep}{3pt}
  \resizebox{1.0\linewidth}{!}{%
  \begin{tabular}{@{}lccc@{}}
    \toprule
    Variant & PSNR\,$\uparrow$ & SSIM\,$\uparrow$ & LPIPS\,$\downarrow$ \\
    \midrule
    (\textit{a}) Unaligned RoPE                     & 18.3132 & 0.6134 & 0.2782 \\
    (\textit{b}) Aligned RoPE, zero offsets         & 18.1341 & 0.6079 & 0.2977 \\
    (\textit{c}) Ours (Stage~1, aligned RoPE + offsets)  & 18.6764 & 0.6335 & 0.2645 \\
    \midrule
    (\textit{d}) Ours (Stages~1+2)                  & \underline{19.0474} & \underline{0.6468} & \textbf{0.2241} \\
    (\textit{e}) Multi-stage relight-only training  & 18.7240 & 0.6441 & 0.2315 \\
    (\textit{f}) Single-stage multi-task training   & 16.9998 & 0.5846 & 0.2921 \\
    (\textit{g}) Ours (Stages~1+2+3, multi tasks)       & \textbf{19.2039} & \textbf{0.6545} & \underline{0.2245} \\
    \bottomrule
  \end{tabular}
  }
\vspace{-1mm}
\end{table}

%% file: sec/5_conclusion.tex
\section{Conclusion}
\label{sec:conclusion}

We present an interactive video relighting system \paper centered on Lightgram, a pixel-aligned illumination representation.
Lightgram supports diverse light types and dynamics, linking real-time shader previews with corresponding lighting conditions for our in-context video generation model.
We equip this model with interleaved RoPE alignment to maintain temporal correspondence and train it through reciprocal illumination learning to strengthen illumination understanding and real-world adaptation.
Together, these designs enable realistic video relighting with accurate lighting control.

\paragraph {Limitations and future directions}
Our lightgrams use screen-space ray tracing (SSRT) to render direct illumination only, limiting visibility estimation for occluded sources and excluding interreflections.
Although our support for varied light types and counts offers workarounds to mimic these effects, plausible bounce lighting still relies on approximate illumination rather than explicit physical cues.
Future work could extend lightgrams to capture richer light transport.

%% file: sec/X_suppl.tex
\clearpage
\setcounter{page}{1}
\maketitlesupplementary

\makeatletter
\renewcommand{\theHsection}{arabicsection.\thesection}
\renewcommand \thesection{S\@arabic\c@section}
\setcounter{section}{0}
\renewcommand \thetable{S\@arabic\c@table}
\setcounter{table}{0}
\renewcommand \thefigure{S\@arabic\c@figure}
\setcounter{figure}{0}
\renewcommand \theequation{S\@arabic\c@equation}
\setcounter{equation}{0}
\makeatother

\section{Design Rationale}
\label{sec:claims}

A central design of \paper is to bridge interactive lighting design and video generation through a shared representation.
Specifically, our proxy renderer makes lighting configurations immediately visible and encodes them as lightgrams for generative control, while also facilitating real-world data construction.
Below, we first detail how the underlying linear response serves both preview and conditioning, and then explain how synthetic lighting supervision and real-world restoration jointly yield controllable, realistic results.

\subsection{A Shared Lighting Interface for Preview and Generation}

Given a lighting configuration $\Lambda$, the proxy renderer computes a linear RGB response $R_\Lambda$ on a standardized material using the estimated scene geometry.
This response combines the contributions of the specified lights, including their intensities, colors, and approximate visibility.
Its values can exceed unity.
We encode this response as a lightgram $L$ through the mapping $\mathcal{T}$:
\begin{equation}
    L = \mathcal{T}(R_\Lambda)
      = \Gamma\left(
          \frac{R_\Lambda}{1 + \max_c R_{\Lambda,c}}
        \right),
    \label{eq:lightgram_supp}
\end{equation}
where $c\in\{r,g,b\}$ indexes RGB channels, $\Gamma$ denotes linear-to-sRGB conversion, and the channel maximum is computed independently at each pixel.
This encoding preserves the spatial and chromatic organization of the rendered response within the bounded range $[0,1]$.

For interactive preview, let $I$ denote an sRGB source frame and $I_{\mathrm{lin}}$ its linear RGB representation.
We composite the linear response $R_\Lambda$ directly with the source to obtain the linear RGB preview $P_{\mathrm{lin}}$:
\begin{equation}
    P_{\mathrm{lin}}
    =
    2^e I_{\mathrm{lin}} \odot (a\mathbf{1} + R_\Lambda),
    \qquad
    I_{\mathrm{lin}} = \Gamma^{-1}(I),
    \label{eq:lightgram_on_source}
\end{equation}
where $a$ is a scalar controlling source-image retention, $e$ is exposure in EV, $\mathbf{1}=(1,1,1)$ is the all-ones RGB vector, and $\odot$ denotes element-wise multiplication.
Optional highlight compression and sRGB conversion produce the displayed preview.
The effective gain $2^e(a\mathbf{1}+R_\Lambda)$ determines the local photometric change, producing brightening where it exceeds unity and attenuation where it falls below unity, evaluated per channel.
The bounded lightgram and the linear response used for compositing are thus two representations of the same lighting configuration with different roles.

Applying the response to the captured image retains the influence of its original illumination.
Under the simplified model $I_{\mathrm{lin}}\approx A\odot S_{\mathrm{src}}$, where $A$ denotes diffuse albedo and $S_{\mathrm{src}}$ denotes shading under the captured illumination, the preview contains $A\odot S_{\mathrm{src}}\odot(a\mathbf{1}+R_\Lambda)$, up to exposure scaling.
Existing shading therefore remains influential, and black pixels remain black.

Within this formulation, users can rapidly inspect light placement, color, spatial extent, and proxy-predicted shadow directions.
Furthermore, time-varying lighting trajectories can be flexibly authored by specifying sparse keyframes with automatic property interpolation, as demonstrated in \cref{fig:shader} of the main paper.
The corresponding lightgram then carries the configured lighting into the generation process.

In summary, the proxy renderer provides immediate feedback for lighting design, while the generative model takes the resulting lightgram as approximate cues and translates them into the target lighting edits, synthesizing realistic appearance beyond the coarse preview.

\subsection{Real-World Adaptation with Complementary Supervision}

Learning the conditional mapping from lightgram cues to target lighting edits fundamentally relies on synthetic pairs featuring known lighting interventions alongside aligned scene content and motion, which provide direct supervision for illumination replacement, extraction, addition, and subtraction.
Consequently, the synthetic training stages teach the model how to interpret lightgram cues across distinct editing operations.

To extend this conditional editing capability to real-world videos while preserving natural appearance, Stage~3 combines continued synthetic supervision with real-world photometric restoration.
Specifically, we employ the same proxy shader as the real-time previewer to perturb the original video $V_t$, generating a degraded input $V_s$ with varying brightening, attenuation, and color shifts, while retaining $V_t$ as the reconstruction target.
To ensure effective supervision, we filter out samples exhibiting negligible appearance changes or overly dark renders.

For real-world illumination replacement, restoration is conditioned on the original lightgram $L_t$ estimated by the Stage~1 model, whereas for subtraction, it is conditioned on the perturbation lightgram $L_\Delta$ associated with the applied shading change.
Within the real-world task mixture, we place primary emphasis on illumination replacement, using perturbation-conditioned subtraction as complementary restoration supervision (see \cref{tab:supp_task_sampling} for detailed sampling proportions).

This hybrid training strategy assigns distinct, complementary roles to the two data sources: synthetic pairs preserve precise supervision over the intended lighting operations, while real-world photometric restoration anchors the model to photorealistic video appearance through genuine reconstruction targets. Consequently, the model adapts effectively to real-world domains without sacrificing the fine-grained lighting controllability learned from rendered data.

Our ablation studies confirm this complementarity. As shown in \cref{fig:ablation} of the main paper, Stage~3 improves shading naturalness, eliminating the pronounced CG-like appearance exhibited by the synthetic-only baseline. Meanwhile, \cref{tab:ablation} demonstrates that quantitative performance on synthetic benchmarks is largely preserved. Together, these results validate the synergy of combining aligned synthetic control supervision with real-world appearance adaptation.

\section{Implementation Details}
\label{sec:implementation_details}

\subsection{Model}

\paragraph{Backbone and adaptation}
We initialize our model from the pretrained MiniMax H3~\cite{h3} reference-conditioned backbone, which contains 50 Transformer blocks with hidden dimension $D=5376$ and two semantic token-refinement blocks.
RGB videos and lightgrams share the pretrained video VAE, while Qwen3-VL provides the semantic context, as described in \cref{sec:relighting_model}.
Both encoders remain frozen, and their outputs are cached for training.
We follow the VAE's temporal grid: $17n+5$ input frames yield $5n+2$ latent frames.
The VAE produces 24-channel latents with $16\times$ spatial compression, which are tokenized into $1\times2\times2$ spatiotemporal patches. 
In the Transformer backbone, each attention module uses 56 heads of dimension 128, giving an attention width $A=7168$.
We apply LoRA~\cite{lora} jointly to the query--key--value projections ($D\rightarrow3A$) and to the attention output projection ($A\rightarrow D$) in all 50 Transformer blocks and both token-refinement blocks.
We set the LoRA rank and scaling parameter to $r=\alpha=256$.

\paragraph{Task-specific modulation}
For each Transformer block $\ell$, task $q$ has an independent, zero-initialized residual embedding $b_{\ell,q}\in\mathbb{R}^{6D}$.
It is added to the video-modality modulation parameters after projecting the timestep embedding:
\begin{equation}
  m_{\ell,q}(\tau)=m_{\ell}^{\mathrm{base}}(\tau)+b_{\ell,q}.
  \label{eq:supp_task_modulation}
\end{equation}
The six $D$-dimensional components control the shift, scale, and residual gate of the attention and feed-forward branches.
A separate zero-initialized task embedding in $\mathbb{R}^{2D}$ adjusts the shift and scale of the final adaptive normalization layer.
These AdaLN-style modulations operate on the backbone's RMS-normalized features.
The task identity is shared across the conditioning and output video tokens, while each stream retains its own noise-level conditioning.

\paragraph{Interleaved RoPE alignment}
We retain the backbone's native temporal grid, whose successive spacings repeat the pattern $\frac{5}{3}(1,4,4,4,4)$.
These coordinates use a clock of 40 positional units per second at 24\,fps, giving $\Delta_{\min}=5/3$ in \cref{eq:alignment}.
For the RGB-output tasks, we use
\begin{equation}
  (\delta_{\mathrm{source}},\delta_{\mathrm{lightgram}},
  \delta_{\mathrm{target}})
  =\left(\frac{5}{6},-\frac{5}{12},0\right).
  \label{eq:supp_rope_offsets}
\end{equation}
The offset span is therefore $5/4<5/3$.
All streams share the spatial coordinate grid.
For illumination extraction, the conditioning video occupies the target RGB slot at zero offset, and the predicted lightgram retains its offset of $-5/12$, preserving the relative positions of these two slots.

\subsection{Training and Inference}

\paragraph{Flow-matching objective}
Using the convention $Z^{\tau}=\tau Z+(1-\tau)\epsilon$ in \cref{eq:flow_matching}, we sample a uniform variable $u\sim\mathcal{U}(0,1)$ and apply the video noise shift
\begin{equation}
  \sigma=\frac{12u}{1+11u},\qquad \tau=1-\sigma.
  \label{eq:supp_noise_sampling}
\end{equation}
We use a Gaussian-shaped loss weight with a constant floor:
\begin{equation}
  w(\tau)=0.1+0.9\,
  \frac{\exp\!\left[-2(\tau-0.5)^2\right]-\exp(-0.5)} {1-\exp(-0.5)}.
  \label{eq:supp_loss_weight}
\end{equation}
This weight reaches one at $\tau=0.5$ and remains at least $0.1$ throughout the interval.
Conditioning video latents use a fixed flow time $\tau_{\mathrm{ref}}=0.999$, giving $Z_{\mathrm{ref}}^{\mathrm{cond}}=0.999Z_{\mathrm{ref}}+0.001\epsilon_{\mathrm{ref}}$.
Supervision is applied to the selected output block: the lightgram for extraction and the RGB video for the other tasks.

\paragraph{Training curriculum}
Following the three-stage curriculum introduced in \cref{sec:reciprocal_learning}, we progressively expand the training tasks and incorporate real-world supervision.
Each training sample comprises two aligned RGB videos under different illumination and their associated lightgrams, and can support multiple tasks.
Stage 1 (illumination grounding) uses roughly 24k video pairs, with around 75\% object-level data rendered from Objaverse~\cite{objaverse} assets in Blender and around 25\% scene-level data rendered in Unreal Engine.
These pairs provide balanced supervision for relighting and illumination extraction, grounding the model's understanding of the correspondence between lighting and appearance.
Stage 2 (incremental lighting control) uses roughly 10k pairs drawn from Objaverse to introduce addition and subtraction while continuing relighting and extraction.
Stage 3 (real-world adaptation) uses roughly 8k pairs, incorporating real-world relighting and subtraction while retaining synthetic supervision for all four tasks.
\Cref{tab:supp_training} summarizes the data scales and training schedules.

At each stage, we uniformly sample a video pair from the training set and then select one of its supported tasks using stage-specific sampling weights.
The expected sampling proportions in \cref{tab:supp_task_sampling} account for both the stage-specific data composition and task sampling weights.
For real-world relighting, we use precomputed predictions from illumination extraction, with the textured area-light approximation described in \cref{sec:data_engine} as a fallback.
Real-world subtraction uses the known perturbation lightgram, and both real-world tasks reconstruct the original videos.

\begin{table}[t]
  \caption{\textbf{Three-stage training configuration.}}
  \label{tab:supp_training}
  \centering
  \small
  \renewcommand{\arraystretch}{1.1}
  \begin{tabular}{lccc}
    \toprule
    Setting & Stage 1 & Stage 2 & Stage 3 \\
    \midrule
    Video pairs & $\sim$24k & $\sim$10k & $\sim$8k \\
    Training iterations & 1500 & 750 & 500 \\
    Learning rate & $10^{-4}$ & $5\times10^{-5}$ & $5\times10^{-5}$ \\
    \bottomrule
  \end{tabular}
\end{table}

\begin{table}[t]
  \caption{\textbf{Expected task sampling proportions.} Percentages account for the stage-specific data composition and task sampling weights.}
  \label{tab:supp_task_sampling}
  \centering
  \small
  \begin{tabular}{llccc}
    \toprule
    Source & Task & Stage 1 & Stage 2 & Stage 3 \\
    \midrule
    \multirow{4}{*}{Synthetic}
      & Relighting & 50\% & 36\% & 25\% \\
      & Extraction & 50\% & 36\% & 25\% \\
      & Addition & --- & 18\% & 15\% \\
      & Subtraction & --- & 10\% & 10\% \\
    \midrule
    \multirow{2}{*}{Real-world}
      & Relighting & --- & --- & 15\% \\
      & Subtraction & --- & --- & 10\% \\
    \bottomrule
  \end{tabular}
\end{table}

\paragraph{Training}
In our model, trainable parameters comprise the LoRA adapters and the task-specific modulation embeddings, totaling approximately 531M.
We use AdamW~\cite{adamw} with $(\beta_1,\beta_2)=(0.9,0.999)$ and $\epsilon=10^{-8}$, weight decay $0.01$.
The learning rate remains constant within each stage.
Each successive stage starts from the preceding stage's trained weights and initializes fresh optimizer states.

\paragraph{Inference}
Given a source video and a user-specified lightgram, we encode both into latent representations, initialize the target latent with Gaussian noise, and integrate the learned velocity field using the H3 first-order sampler with shift factor 12.
Our default schedule contains 50 noise levels, including the terminal zero, corresponding to 49 model evaluations.
The conditioning streams remain fixed throughout sampling, and only the selected output latent is updated.
The video VAE decodes the result at 24\,fps using the same spatial canvas as the source.

\begin{figure*}[t]
  \centering
  \includegraphics[width=0.95\linewidth]{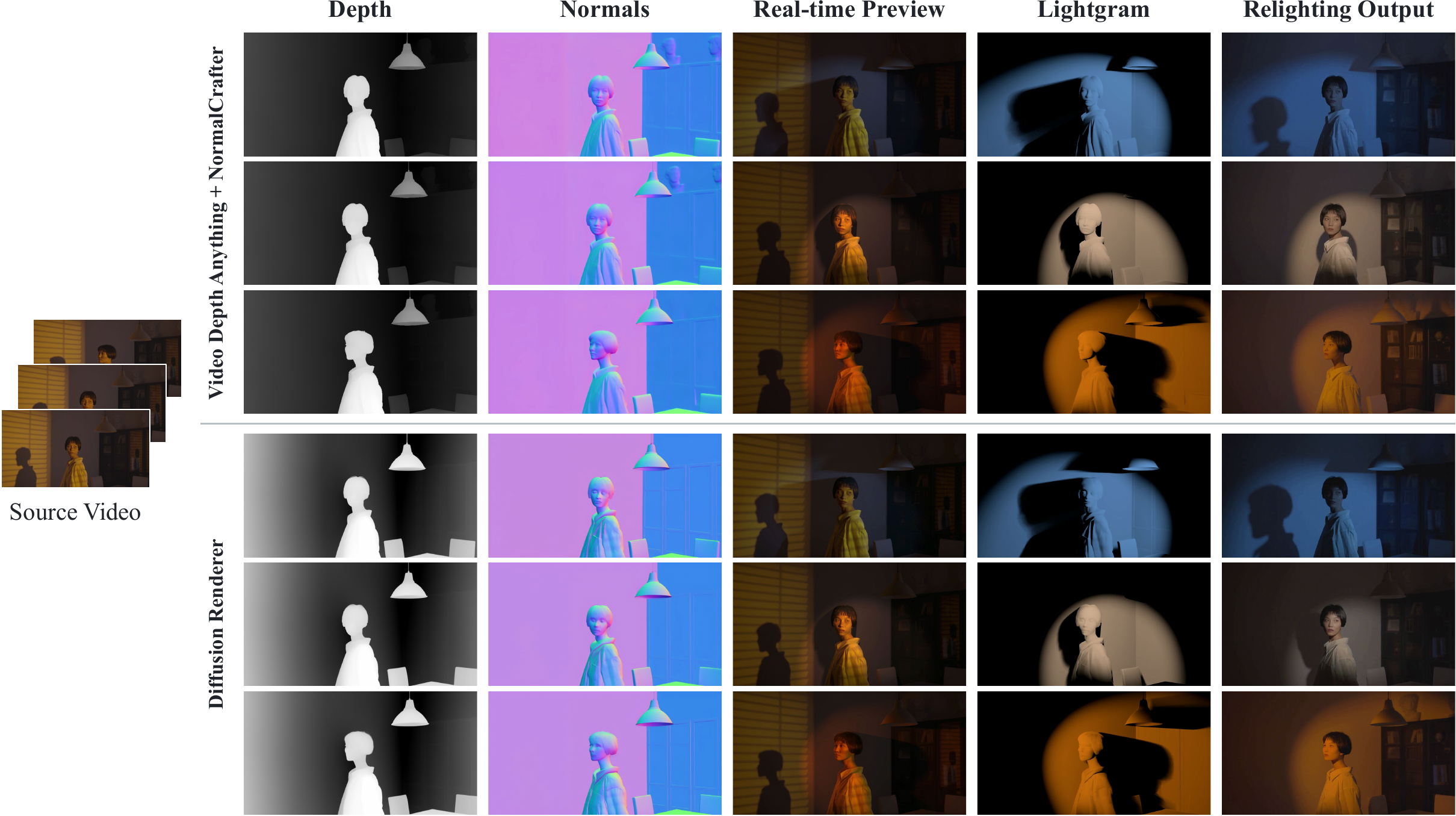}
  \caption{\textbf{Relighting with coarse geometric priors.} Corresponding frames use Video Depth Anything~\cite{vda} and NormalCrafter~\cite{normalcrafter} (upper group) or DiffusionRenderer~\cite{diffusionrenderer} (lower group). We adjust the spotlight's 3D position to account for differences in the estimated depth scales and approximately match its placement and direction relative to the subject. In this example, the final relit videos retain similar overall illumination despite differences in the previews and lightgrams.}
  \label{fig:coarse_prior}
\end{figure*}

\subsection{Data}

\paragraph{Real data preprocessing}
We resample all videos to 24\,fps and use clips of up to 158 frames, approximately 6.6\,s.
Frames are resized with preserved aspect ratio using a nominal short edge of 768 pixels and a pixel budget of $768\times1344$, with spatial dimensions aligned to multiples of 32.
The RGB streams and their lightgrams use corresponding temporal samples and spatial transforms. We estimate monocular normals and depth from videos using NormalCrafter~\cite{normalcrafter} and
Video Depth Anything~\cite{vda}, respectively.

\paragraph{Dataset curation and postprocessing}
As described in \cref{sec:data_engine}, we curate source--target video pairs $(V_s,V_t)$ with aligned lightgrams representing target illumination or incremental lighting components. Rendered lightgrams are produced by our custom proxy shader (\cref{sec:lightgram}), which computes direct-illumination responses on a diffuse screen-space proxy via screen-space ray tracing.

For synthetic data, we render object-level RGB video pairs from 20k Objaverse assets~\cite{objaverse} in Blender and scene-level pairs from 38 curated 3D assets in Unreal Engine.
Both engines provide ground-truth depth and normals.

For real-world data, we retain the in-the-wild original video as the target $V_t$ and synthesize a perturbed source $V_s$. Given a sampled lighting configuration $\Lambda$ and monocularly estimated depth and normals, the proxy shader computes a linear-RGB response $R_\Lambda$, which is mapped to sRGB to form the perturbation lightgram (\cref{eq:lightgram_supp}). We apply $R_\Lambda$ on linearized target following \cref{eq:lightgram_on_source} and convert the result back to sRGB to obtain $V_s$.

We also explored synthesizing pairs with a pretrained inverse-rendering model~\cite{diffusionrenderer} that jointly estimates albedo, depth, and normals. We observed that its predicted G-buffers exhibited temporal flicker and occasional identity degradation, while albedo estimates sometimes retained original shading or reproduced the input frames. These inconsistencies and the high computational cost motivated our use of specialized monocular estimators and the construction of real-world relighting pairs by compositing lightgrams with the original videos in linear RGB.

\begin{figure}[t]
\centering
\includegraphics[width=1.0\linewidth]{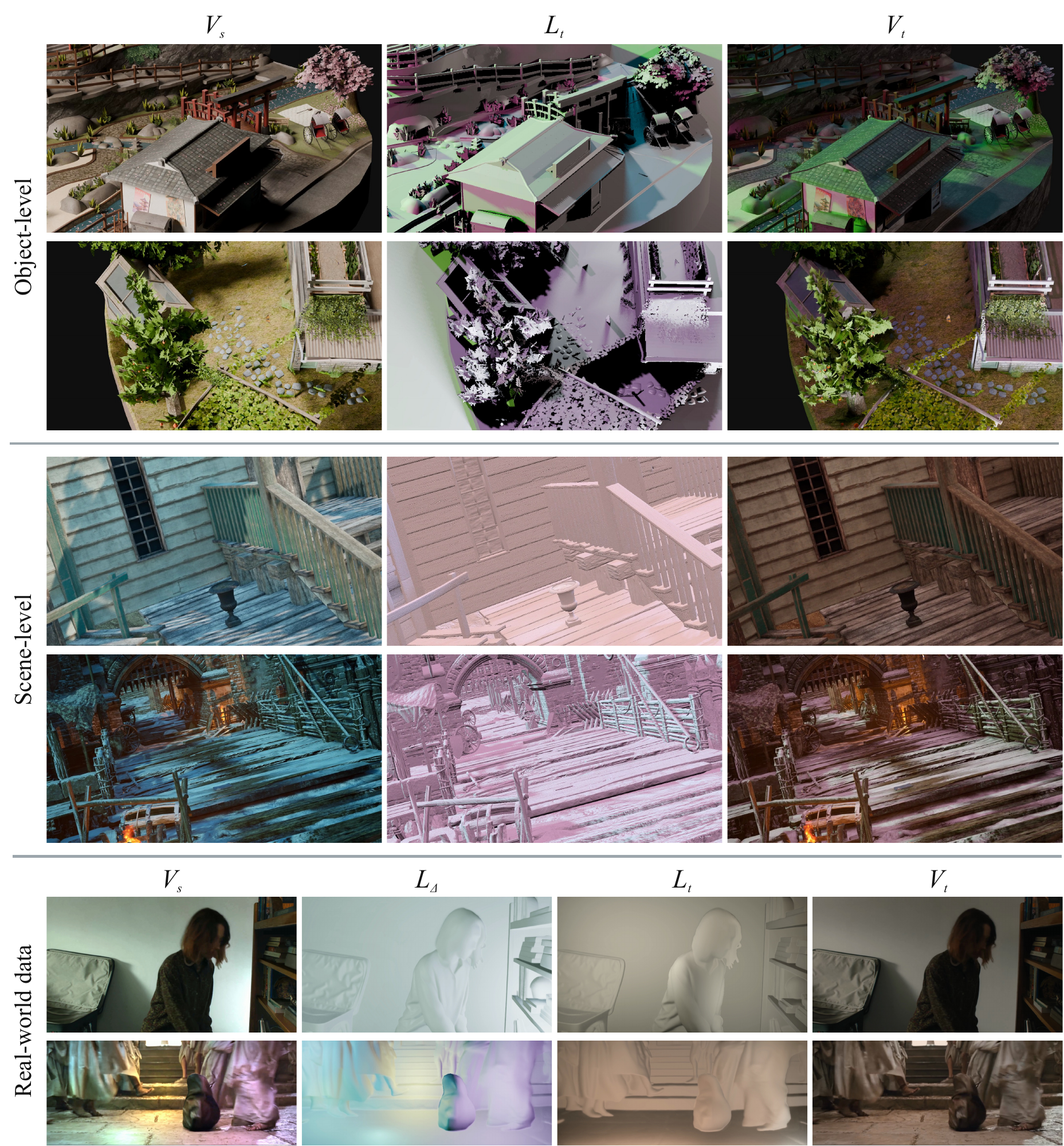}
\caption{\textbf{Training dataset overview.}
We show exemplar synthetic and real-world video pairs with their associated lightgrams.}
\label{fig:supp_dataset}
\end{figure}

Finally, we filter pairs with insufficient first-frame appearance changes using a PSNR threshold, reject excessively dark sequences, and remove samples with low aesthetic scores predicted by a pretrained DOVER++ model~\cite{dover_pp}.
\Cref{fig:supp_dataset} provides an overview of our synthetic and real-world training data.

\paragraph{Structured captioning}
We develop a two-stage captioning pipeline following MiniMax H3's structured prompt format~\cite{h3} for our lighting tasks. Both stages use Gemini-3.8-Flash~\cite{google2026gemini38flash}.

First, we generate \emph{meta captions} describing RGB appearance and lightgram illumination separately for each lighting configuration. The system prompt identifies the lightgram as an approximate rendering of the same subject on white screen-space proxy surfaces. For RGB videos, it directs the model to describe subject appearance, materials, and motion. For lightgrams, it emphasizes light colors, spatial distributions, and temporal changes, distinguishing illumination from material appearance. 
Second, we provide the meta captions and corresponding videos to the model to generate task-specific intermediate representation (IR) captions. Each IR follows H3's six-field structure:

\begin{enumerate}
    \item \textbf{Subject definitions:} Assigns reference tags such as \texttt{<Video 1>} and \texttt{<Video 2>} to the inputs and defines their roles.

    \item \textbf{Summary:} States the intended edit and begins with a task tag:
    \begin{itemize}
        \item {[rgb-conditioned relighting + video editing]}
        \item {[rgb-conditioned illumination addition + video editing]}
        \item {[rgb-conditioned illumination subtraction + video editing]}
        \item {[rgb-conditioned illumination extraction + video editing]}
    \end{itemize}

    \item \textbf{Retention analysis:} Specifies which attributes to preserve or modify from each input.

    \item \textbf{Detailed description:} Integrates these constraints into a description of the desired appearance, lighting, and temporal behavior.

    \item \textbf{Overall soundscape:} Set to \texttt{N/A}.

    \item \textbf{Non-diegetic music:} Set to \texttt{N/A}.
\end{enumerate}

For RGB generation tasks, the instructions emphasize that lightgrams contain only direct illumination and allow plausible indirect illumination in the generated result while preserving the specified lighting intent.

\begin{figure}[t]
    \centering
    \includegraphics[width=\columnwidth]{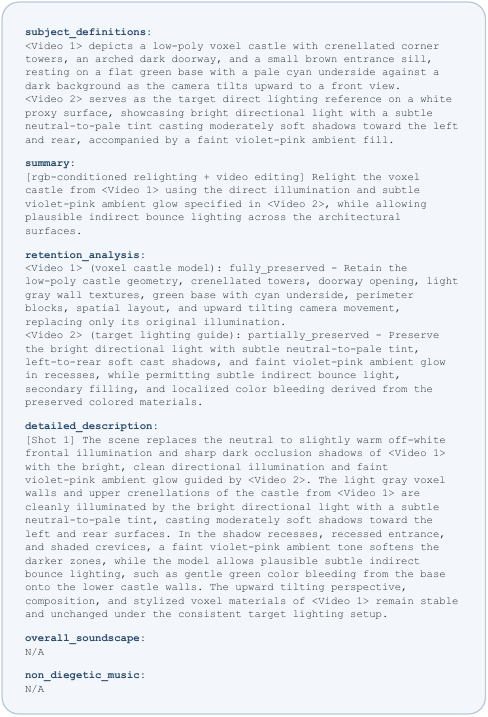}
    \caption{\textbf{Example context IR caption} for illumination replacement. Reference tags distinguish source appearance from target lighting guidance, while structured fields specify the editing intent and preservation requirements.}
    \label{fig:context_ir_example}
\end{figure}

\Cref{fig:context_ir_example} shows an example context IR caption for relighting (illumination replacement). Addition, subtraction, and extraction reuse the per-sample meta captions with task-specific IR instructions.

\paragraph{Test datasets}
Our quantitative synthetic test set comprises 66 samples, including object-level dynamic sequences from the ABO~\cite{abo} and GSO~\cite{gso} datasets, as well as scene-level dynamic data rendered in Unreal Engine.
All evaluated assets and scenes are strictly held out from the training set.

For the VBench evaluation, we additionally collect 50 in-the-wild videos from Pexels~\cite{pexels} and Mixkit~\cite{mixkit}.
Since VBench assesses perceptual and temporal quality without requiring ground-truth relit videos, only target lighting conditions are needed.
We thus construct a pool of point-light configurations and randomly sample from it to pair with each video.
The pool consists of 25 distinct lighting configurations, including static single-light, dynamic single-light, and static multi-light setups with up to three distinct point light sources.

\section{Lighting Accuracy Evaluation Details}
\label{sec:supp_lighting_evaluation}

Our extraction-based protocol provides an auxiliary measure of lighting accuracy in lightgram space, complementing the RGB-domain metrics in \cref{sec:exp_settings} and qualitative comparisons in \cref{fig:qualitative_baselines}.
The underlying motivation is that RGB similarity jointly entangles scene content, texture, geometric alignment, and illumination.
Consequently, merely copying the source video can still achieve deceptively high RGB similarity on certain samples, despite failing to apply the requested lighting change.

Specifically, we freeze our final model and use its extraction mode $\mathcal{E}$ for all evaluated methods.
Given a generated relighting video $\widehat{V}$, we extract $\widehat{L}=\mathcal{E}(\widehat{V})$ and compare it with the renderer-produced target lightgram $L_t$ encoding the requested illumination.
The extractor and target conditions are shared across methods. The scores quantify agreement under this fixed learned mapping.

\paragraph{Median RGB angular error}
Following the recovery angular error used in illuminant estimation~\cite{finlayson2017reproduction}, we measure the angle between corresponding RGB vectors.
For pixels $p$, the error in degrees is
\begin{equation}
  \theta_p = \frac{180}{\pi}\arccos\!\left(
    \frac{\widehat{L}_p^\top L_{t,p}}
    {\|\widehat{L}_p\|_2\|L_{t,p}\|_2}
  \right).
  \label{eq:lighting_angular_error}
\end{equation}
We average pixels within each frame, average the resulting frame errors within each video, and report the median of these video-level means.
This measure is invariant to positive intensity scaling and captures agreement in light-color direction.

\paragraph{Mean SSIM}
We compute SSIM~\cite{ssim} between extracted and target lightgrams for each frame, average over frames within each video, and then average over videos.
SSIM assesses luminance, contrast, and local spatial structure, complementing angular error's insensitivity to intensity scaling.
Together, the two metrics characterize light-color agreement and the spatial distribution of lighting responses.

To contextualize the extractor's residual error, we also apply it to ground-truth target RGB videos, yielding the GT-extracted reference in \cref{tab:lighting_accuracy}.

\section{Independent Shading Evaluation}
\label{sec:shading_evaluation}

To complement our extraction-based evaluation (\cref{sec:supp_lighting_evaluation}), we use Marigold-IID-Lighting v1.1~\cite{ke2025marigold}, a third-party intrinsic decomposition model trained independently of our framework.
We use its official pretrained weights, freeze the estimator, and apply it identically to the generated and ground-truth RGB frames on the same synthetic test set as the main paper (\cref{sec:exp_settings}).
Marigold separates albedo, colored diffuse shading, and non-diffuse residuals.
We compare its shading estimates to assess illumination color and spatial lighting patterns.

\paragraph{Evaluation metrics}
We follow the metric definitions and aggregation rules in \cref{sec:supp_lighting_evaluation}, applying median RGB angular error and mean SSIM to predicted and reference shading.

Alongside SSIM at the estimator's native scale, we report scale-aligned SSIM (SA-SSIM) to account for the scale ambiguity of intrinsic decomposition.
SA-SSIM applies one nonnegative least-squares gain per video before scoring:
\begin{equation}
    \alpha_i =
    \frac{\langle \widehat{S}_i, S_i^{\mathrm{ref}}\rangle}
    {\max\!\left(\|\widehat{S}_i\|_2^2, 10^{-12}\right)}.
    \label{eq:shading_scale}
\end{equation}
Here, $\widehat{S}_i$ and $S_i^{\mathrm{ref}}$ stack all evaluated frames, pixels, and RGB channels of the predicted and reference shading for video $i$.
We apply $\alpha_i$ directly to the floating-point predictions for scoring. This shared gain preserves chromaticity and relative spatial and temporal brightness.
Both SSIM variants use a fixed data range of one.

\begin{table}[t]
    \centering
    \small
    \caption{\textbf{Independent shading evaluation} on the same synthetic test set as the main paper, using a frozen Marigold estimator. SA-SSIM aligns one gain per video. Best and second-best results are shown in \textbf{bold} and \underline{underlined}, respectively.}
    \label{tab:independent_shading}
    \resizebox{1.0\linewidth}{!}{
    \begin{tabular}{lccc}
        \toprule
        Method & Median AE ($^\circ$)~$\downarrow$ & SSIM~$\uparrow$ & SA-SSIM~$\uparrow$ \\
        \midrule
        V-RGBX~\cite{vrgbx} & 7.225 & \underline{0.6283} & \underline{0.7231} \\
        LightCtrl~\cite{lightctrl} & 12.427 & 0.3745 & 0.3830 \\
        Relit-LiVE~\cite{relitlive} & 9.275 & 0.5230 & 0.5678 \\
        MiniMax H3~\cite{h3} & \underline{6.954} & 0.6095 & 0.6566 \\
        \midrule
        \textbf{Ours} & \textbf{5.171} & \textbf{0.7057} & \textbf{0.7437} \\
        \bottomrule
    \end{tabular}
    }
\end{table}

\paragraph{Results}
\Cref{tab:independent_shading} shows that our method achieves the lowest median angular error and the highest SSIM under both scale treatments.
Its advantage after video-level gain alignment supports closer agreement in relative shading patterns.
These results are consistent with the overall trend of our extraction-based evaluation (\cref{tab:lighting_accuracy}), providing complementary evidence of lighting accuracy.

\begin{figure*}[t]
  \centering
  \includegraphics[width=0.9\linewidth]{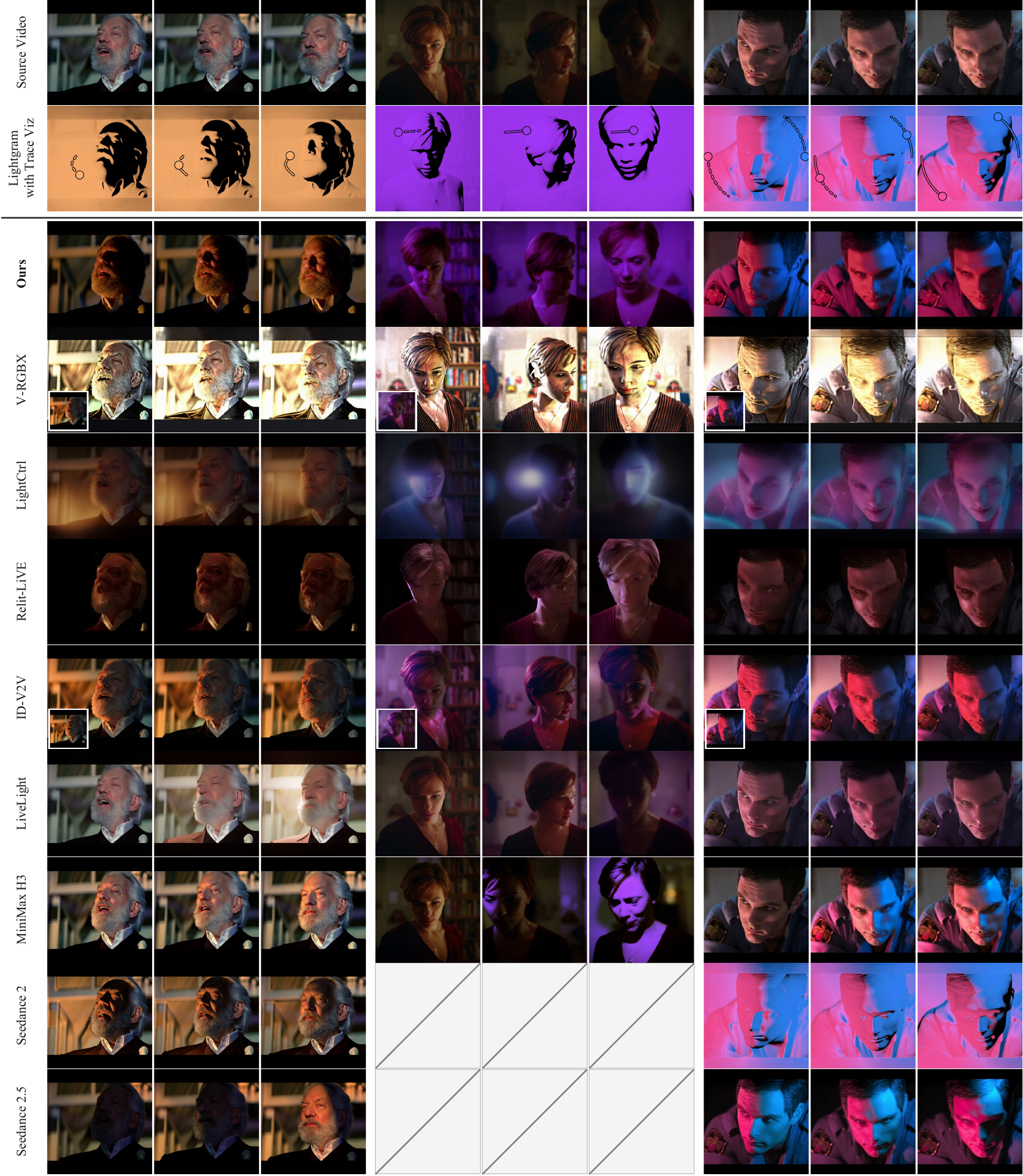}
  \caption{
  \textbf{Additional qualitative comparison.}
We use point lights to align with the baselines' lighting representations for a fair comparison.
V-RGBX~\cite{vrgbx} and ID-V2V~\cite{idv2v} are evaluated under static lighting, as they take only a relit first frame as lighting guidance.
Seedance outputs are omitted for cases blocked by its built-in safety filtering for videos depicting real people.
MiniMax H3 exhibits temporal misalignment with the source video in Case~2 and delayed relighting in Case~3, where the target illumination appears only in later frames.
  }
  \label{fig:suppl_qualitative_baselines_1}
\end{figure*}

\begin{figure*}[t]
  \centering
  \includegraphics[width=0.9\linewidth]{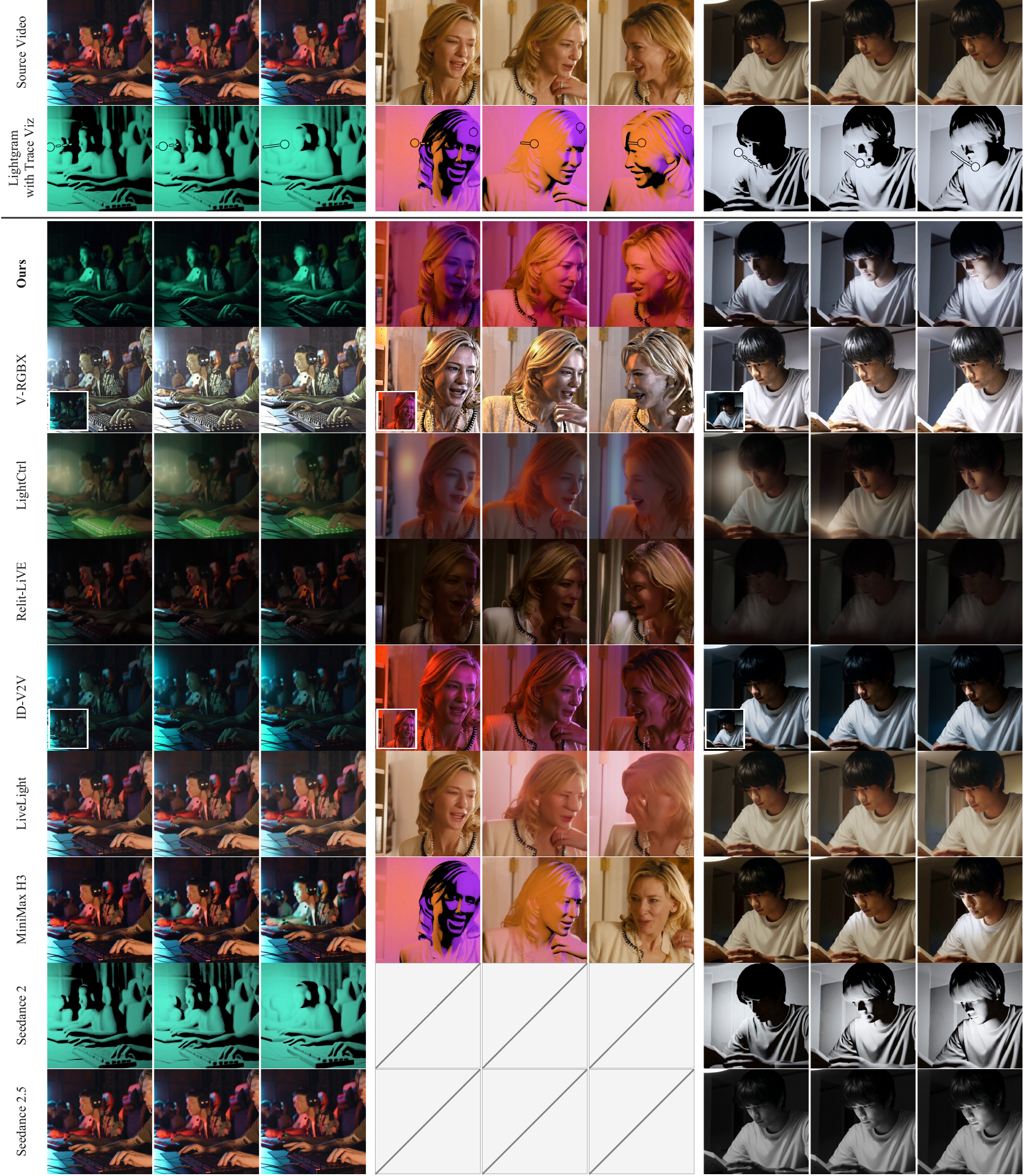}
  \caption{
  \textbf{Additional qualitative comparison } (continued from \cref{fig:suppl_qualitative_baselines_1}).
  We use point lights to align with the baselines' lighting representations for a fair comparison.
  V-RGBX~\cite{vrgbx} and ID-V2V~\cite{idv2v} are shown under static lighting, as they take only a relit first frame as input.
  Outputs for Seedance are omitted on certain cases due to its built-in safety filtering on real-human videos. LiveLight exhibits pronounced facial deformation in the middle case, while Seedance introduces a black-and-white film style in the rightmost case.
  }
  \label{fig:suppl_qualitative_baselines_2}
\end{figure*}

\begin{figure*}[t]
  \centering
  \includegraphics[width=0.9\linewidth]{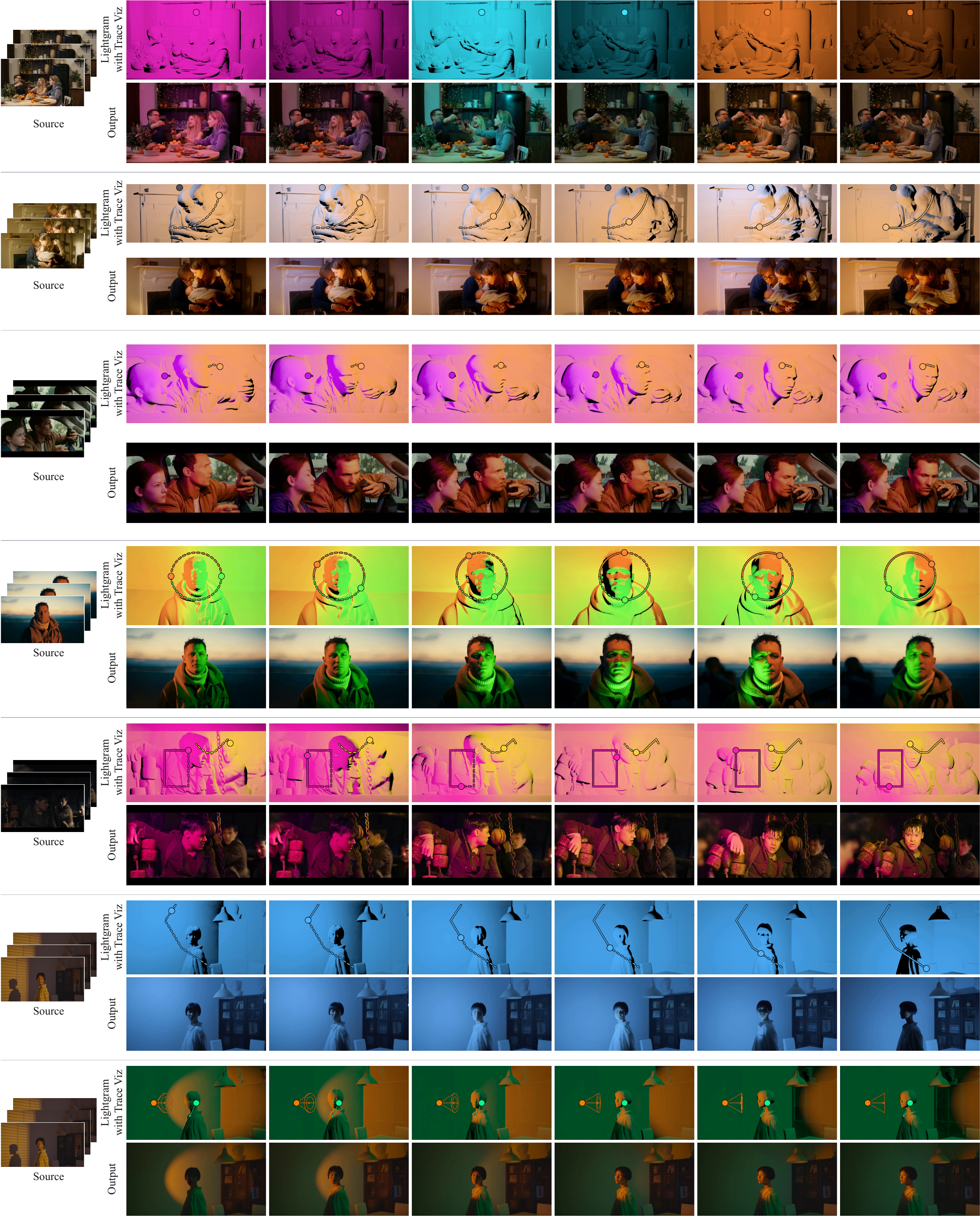}
  \caption{
  \textbf{Additional relighting results of our method on in-the-wild videos.}
  Cases 1--2 demonstrate lighting-intensity changes. Cases 3--6 illustrate spatially varying illumination. Cases 6--7 further show plausible removal of the original warm orange reflection on the bookshelf, unveiling the underlying shelf details with new illumination and shadows.
  }
  \label{fig:suppl_ours_solo}
\end{figure*}

\section{More Experiment Results}

This section presents additional qualitative results and baseline comparisons, including evaluations under environment-map illumination and experiments with coarse geometric priors.
Please refer to the supplementary video for additional dynamic relighting results and baseline comparisons.

\subsection{Qualitative Comparisons with Baselines}

\Cref{fig:suppl_qualitative_baselines_1,fig:suppl_qualitative_baselines_2} extend the qualitative comparison in \cref{fig:qualitative_baselines} with additional real-world examples under the same point-light evaluation protocol.
For some inputs depicting real people, the safety filtering in the evaluated Seedance versions~\cite{seedance2,seedance25} prevents generation; the corresponding outputs are therefore omitted from the figures.

Among the generated results, pretrained MiniMax H3~\cite{h3} exhibits temporal inconsistencies and occasionally blends the source appearance with the lightgram, introducing the proxy's appearance into the relit frames.
Seedance also alters scene appearance beyond the requested lighting change: in a neutral-light example, its output becomes nearly monochromatic, suppressing colors present in the source video.
LiveLight~\cite{livelight} exhibits pronounced facial deformation in the shown comparison, compromising the subject's recognizable identity.

In these comparisons, \paper faithfully follows the requested illumination while preserving facial identity, scene appearance, and coherent motion.

\subsection{More Visualizations}

We present additional qualitative relighting results of our method on diverse in-the-wild videos in \cref{fig:suppl_ours_solo}.

Across varying subjects and target lighting conditions, our model consistently produces realistic shading and coherent temporal dynamics while preserving fine scene details.

\subsection{Environment-map Illumination}

\begin{figure*}[t]
  \centering
  \includegraphics[width=0.9\linewidth]{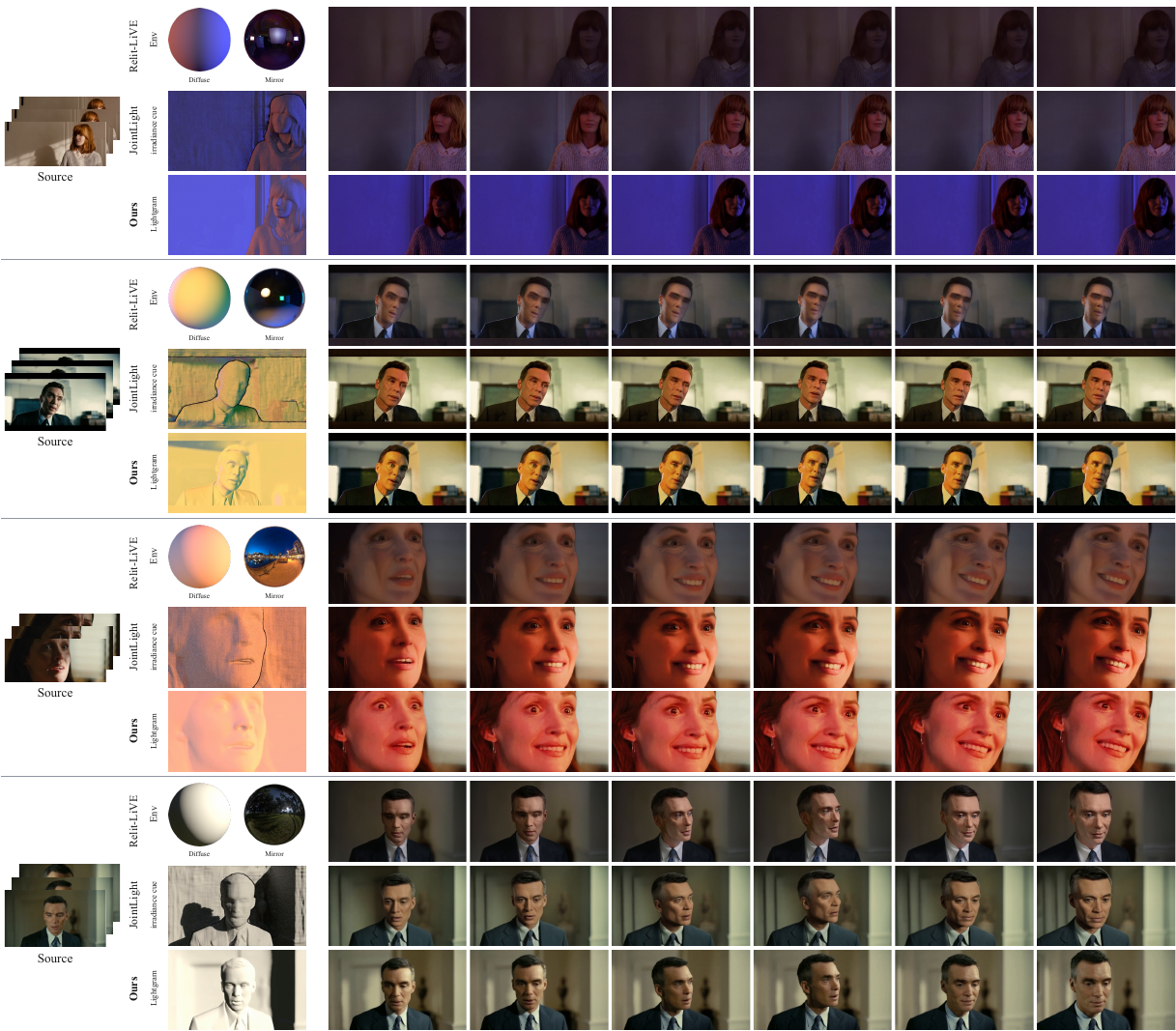}
  \caption{\textbf{Qualitative comparisons under environment-map illumination.}
  We compare our method against environment-map-based baselines, including Relit-LiVE~\cite{relitlive} (evaluated in its native environment-map setting) and the concurrent work JointLight~\cite{jointlight}. Alongside the source and relit frames, we also visualize the actual lighting conditioning representations derived from the target environment map for each method: the raw environment map for Relit-LiVE, 3D-reconstruction-based irradiance cues for JointLight, and our lightgrams.
  }
  \label{fig:suppl_envmap}
\end{figure*}

In \cref{sec:compare} of the main paper, we adapt Relit-LiVE~\cite{relitlive} to a point-light setting for a consistent comparison across baselines.
Here, we further evaluate an environment-map setting that is more favorable to baselines natively operating under environment-map illumination, and additionally include JointLight~\cite{jointlight}, a concurrent work in this setting.
All environment maps are sourced from Poly Haven.
Note that our model is never trained on environment-map data; thus, this evaluation represents a zero-shot generalization test to an out-of-distribution lighting type.

\Cref{fig:suppl_envmap} presents the relit outputs and corresponding lighting representations, using indoor environment maps in Cases~1--2 and outdoor environment maps in Cases~3--4.
Our method produces more expressive lighting changes while preserving the subjects' identities and subtle facial expressions.
Relit-LiVE exhibits pronounced artifacts and identity shifts, weakening subtle facial expressions that are central to acting.
JointLight exhibits stronger skin smoothing and flatter, more ambient illumination, which we hypothesize stems from training with predominantly uniform environment lighting.
Its outputs also have a synthetic, glossy appearance, potentially reflecting its exclusive use of synthetic training data.

\subsection{Relighting with Coarse Geometric Priors}
\label{sec:supp_coarse_prior}

To examine the effect of geometric prior quality in a qualitative case study, we compare our default priors from Video Depth Anything~\cite{vda} and NormalCrafter~\cite{normalcrafter} with depth and normal estimates from DiffusionRenderer~\cite{diffusionrenderer} on the same source video.
In this example, DiffusionRenderer produces coarser depth and normal estimates.
To account for differences in the estimated depth scales, we slightly adjust the spotlight's 3D position to approximately match its placement and illumination direction relative to the subject.
For each set of priors, we render the corresponding preview and lightgram and use the lightgram to condition the same relighting model.

\Cref{fig:coarse_prior} shows how differences in the geometric priors propagate through this process.
The coarser DiffusionRenderer estimates produce less stable shading in the previews and lightgrams, particularly on the face, where frame-to-frame variations in the estimated normals affect the rendered response.
The changed geometric priors also alter shadow shape and placement under the approximately matched lighting configurations.

These differences are less pronounced in the final relit videos, which retain similar overall illumination and facial appearance across the two sets of priors, with some variation in shadow details.
This example suggests that the model can accommodate these local lightgram imperfections while preserving the intended lighting appearance.

\section{Additional Ablation Details}
\label{sec:supp_ablation}

These settings complement the ablation analysis in \cref{sec:ablation} and the comparisons in \cref{tab:ablation,fig:ablation}.

\paragraph{Controlled comparisons}
For (\textit{a})--(\textit{c}), we hold the Stage~1 data, tasks, and training budget fixed and vary the positional encoding.
All remaining variants use the proposed alignment with role-specific offsets.
The relight-only variant (\textit{e}) removes extraction, addition, and subtraction tasks, with the same data mixtures and training settings as the corresponding stages of (\textit{g}).

\paragraph{Curriculum and lighting diversity}
For single-stage mixed training (\textit{f}), we use the same training pairs and tasks as (\textit{g}), with all data sources and tasks available from the outset.
We also use fixed sampling weights to match the expected joint proportions of data sources and tasks accumulated over the full curriculum of (\textit{g}).
Real-world training targets predominantly contain relatively flat natural illumination, with fewer strongly stylized lighting examples.
Therefore, their early introduction can bias optimization toward subdued lighting responses, consistent with the muted colors of (\textit{f}) in \cref{fig:ablation}.
In contrast, our curriculum first establishes diverse lighting responses on synthetic data and then adapts to real-world appearance.

\section{Failure Cases}
\label{sec:failure}

\begin{figure}[t]
  \centering
  \includegraphics[width=\linewidth]{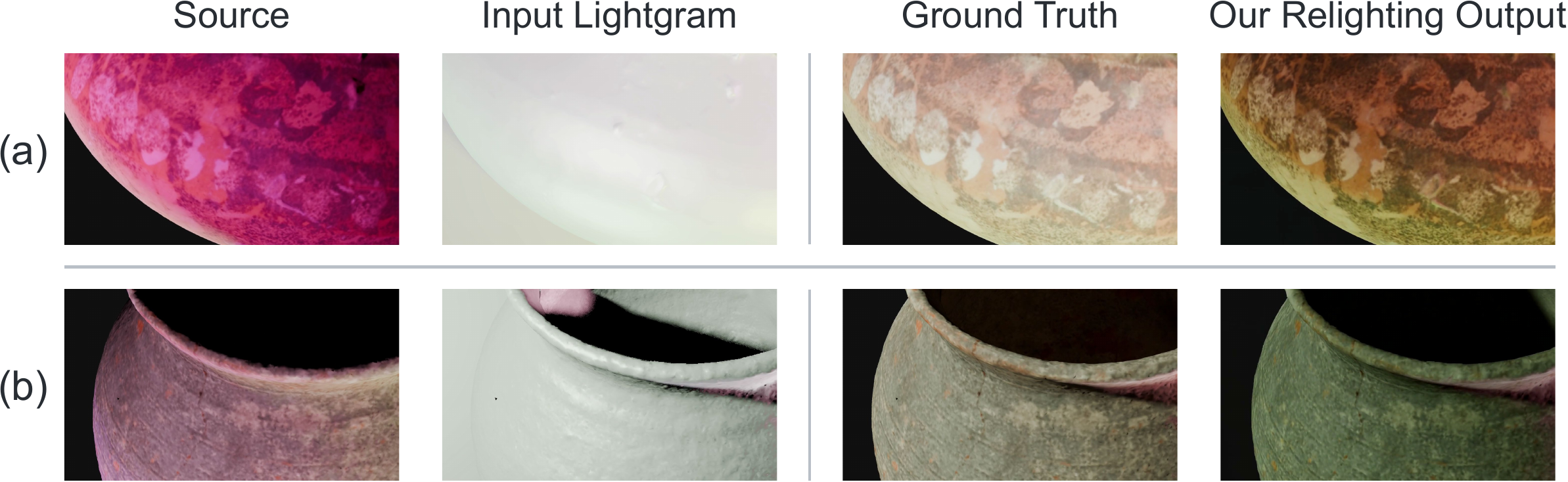}
  \caption{\textbf{Material--illumination ambiguity under colored source lighting.} Our lighting replacement correctly preserves structure and texture but produces color shifts relative to the ground truth.}
  \label{fig:failure}
\end{figure}

Lighting replacement relies on the source video to preserve material appearance and on the target lightgram to specify the new illumination.
However, under colored source lighting, it can be unclear whether an observed color comes from the material itself, the illumination, or both, potentially resulting in color shifts after relighting.

\Cref{fig:failure} illustrates this ambiguity on weathered pottery, for which a wide range of material colors is plausible.
The model largely preserves structure and texture layout, but produces an overly dark orange-brown appearance in (a) and an excessive olive-green cast in (b).
These color shifts suggest an incorrect interpretation of material color under the original illumination.

In practice, this failure mode primarily affects objects with unconstrained intrinsic colors.
For common real-world subjects carrying strong semantic and albedo priors, such as human, the model usually succeeds in removing colored source lighting and applying the target illumination.
Moreover, to further mitigate this ambiguity during training, we typically construct synthetic relighting pairs with relatively flat, neutral source illumination, as introduced in \cref{sec:data_engine}.